\documentclass[letterpaper, journal, twoside]{ieeeconf}

\IEEEoverridecommandlockouts                              

\usepackage{amsmath,amsfonts}
\usepackage{algorithmic}
\usepackage{array}
\usepackage{textcomp}
\usepackage{stfloats}
\usepackage{url,soul}
\usepackage{comment}
\usepackage{verbatim}
\usepackage{graphicx}
\usepackage{amssymb}

\usepackage{mwe}

\usepackage{cite}

\usepackage[ruled,vlined]{algorithm2e}
\usepackage{color}
\usepackage{xcolor}

\def\BibTeX{{\rm B\kern-.05em{\sc i\kern-.025em b}\kern-.08em
    T\kern-.1667em\lower.7ex\hbox{E}\kern-.125emX}}

\usepackage{balance}

\usepackage[colorlinks,bookmarksopen,bookmarksnumbered,citecolor=black,urlcolor=black,linkcolor=purple]{hyperref}

\newtheorem{rem}{Remark}[section]

\newcommand{\boxend}{\hfill \ensuremath{\blacksquare}}

\newcommand{\real}{{\mathbb{R}}}

\allowdisplaybreaks

\title{\LARGE \bf
Sequential Gaussian Variational Inference for Nonlinear State Estimation and Its Application in Robot Navigation
}

\author{Min-Won Seo$^{1}$ and  Solmaz S. Kia$^{1}$, \emph{Senior Member, IEEE}
\thanks{$^{1}$Min-Won Seo, and Solmaz S. Kia are with the Department of Mechanical and Aerospace Engineering,
        University of California, Irvine, CA 92697, USA,
        {\tt\small \{minwons,kia\}@uci.edu}}
}

\begin{document}


\maketitle

\begin{abstract}
Probabilistic state estimation is essential for robots navigating uncertain environments. Accurately and efficiently managing uncertainty in estimated states is key to robust robotic operation. However, nonlinearities in robotic platforms pose significant challenges that require advanced estimation techniques. Gaussian variational inference (GVI) offers an optimization perspective on the estimation problem, providing analytically tractable solutions and efficiencies derived from the geometry of Gaussian space. We propose a Sequential Gaussian Variational Inference (S-GVI) method to address nonlinearity and provide efficient sequential inference processes. Our approach integrates sequential Bayesian principles into the GVI framework, which are addressed using statistical approximations and gradient updates on the information geometry. Validations through simulations and real-world experiments demonstrate significant improvements in state estimation over the Maximum A Posteriori (MAP) estimation method.
\end{abstract}

\begin{keywords}
Gaussian variational inference (GVI), nonlinear state estimation, nonlinear Bayesian inference, natural gradient descent, and information geometry.
\end{keywords}

\section{Introduction}

Accurate and efficient management of evolving uncertainties in sequential (online) state estimation is crucial for the robust operation of robotic platforms. However, \emph{nonlinearities} in motion and measurement models pose challenges in addressing uncertainty in closed form. To tackle this issue, nonlinear state estimation has been studied to obtain approximate solutions for sequential estimation~\cite{sarkka2023bayesian}. In this work, we present a formulation of nonlinear state estimation as an optimization problem to provide efficient and approximate solutions based on a statistical measure.


Let's consider a discrete-time system model, represented as conditional probability distributions: 
\begin{align}
\label{eq::CPD}
    x_t \sim p(x_t|x_{t-1}),~~~z_t \sim p(z_t|x_t), 
\end{align}
where $x_t \in \real^{n_x}$ and $z_t \in \real^{n_z}$ respectively denote the system's state of interest and the observed data related to the state at time $t \in \mathbb{N}_{>0}$. The model in~\eqref{eq::CPD} represents the state transition between $x_t$ and $x_{t-1}$, and the likelihood function of observing $z_t$ given $x_t$. In the context of robotic systems, these models can be tailored to represent various functional requirements, such as sensor and collision avoidance models, depending on the task~\cite{mukadam2016gaussian, kim2021probabilistic}.

In sequential Bayesian principles, given the models in~\eqref{eq::CPD}, the objective for state estimation is computing posterior probability density function (PDF) $p(x_t|z_{1:t})$ at time $t$ using \emph{Bayes' theorem} and the \emph{law of total probability} from
\begin{align}
\label{eq::BayesRule}
    p(x_t|z_{1:t}) \!=\! \frac{p(z_t|x_t)\!\int\! p(x_t|x_{t-1}) 
    p(x_{t-1}|z_{1:t-1}) \text{d}x_{t-1}}{\int p(z_t|x_t)p(x_t|z_{1:t-1}) \text{d}x_t}.
\end{align}

Solving equation \eqref{eq::BayesRule} presents mathematical challenges due to \emph{integral terms} arising from the Chapman-Kolmogorov equation~\cite{papoulis2002probability} and the normalization term. A primary challenge in handling nonlinear models is the infeasibility of finding a closed-form (global) solution to \eqref{eq::BayesRule}, even under Gaussian noise assumptions. Therefore, in the presence of \emph{nonlinearities} in $p(x_t|x_{t-1})$ and $p(z_t|x_t)$, there is a need to explore nonlinear approximate state estimation techniques with a lower deviation from an optimal solution. This work aims to address this intricate challenge by proposing a variational inference approach.

There are two primary approaches for nonlinear state estimation: one is based on global and the other is based on local approximations~\cite{arasaratnam2007discrete, sarkka2023bayesian}. Global approximations involve methods based on finite sample space, such as the Hidden Markov Model~\cite{svensen2007pattern}, the Sequential Monte Carlo~\cite{doucet2001introduction}, and the Markov Chain Monte Carlo~\cite{gilks1995markov}. On the other hand, local approximation relies on the parametric space, particularly employing Gaussian parameters. This includes linearizing nonlinear functions through Taylor expansion, as seen in the Extended Kalman Filter (EKF)~\cite{gelb1974applied}, employing statistical approximations based on the unscented transform and moment matching~\cite{julier1997new}, and using the Laplace approximation~\cite{kass1991laplace}.

Local approximation methods are among the most popular approaches for nonlinear state estimation due to their lower computational cost for online estimation. A key feature of these methods is the determination of parameters locally for a \emph{prespecified} approximated distribution of the posterior probability. 
 To improve the accuracy and efficiency of nonlinear state estimation, this paper leverages inference theory from statistics, particularly variational inference (VI), to offer the flexibility needed to find the best approximate fit for the posterior probability density function (PDF) across the entire parameter space, rather than limiting the focus to local approximations.


VI provides a powerful framework for approximating complex posterior PDFs through optimization rather than sampling, making it highly relevant for state estimation in nonlinear systems.
In the context of VI, solving the state estimation at time $t$ involves using a proposal distribution $q(x_t)$ to approximate $p(x_t|z_{1:t})$ within an optimization framework. The relationship between $q(x_t)$ and $p(x_t|z_{1:t})$ is characterized using the (left) Kullback-Leibler divergence (KLD)~\cite{kullback1951information}: 
\begin{align}
\label{eq::KLD}
    \text{KL}[q(x_t)\|p(x_t|z_{1:t})]  =\!\! \int_{-\infty}^{\infty} \!\! q(x_t) \ln{ \frac{q(x_t)}{p(x_t|z_{1:t})} } \text{d}x_t \geq 0,\!
\end{align}
where the equality holds when $q(x_t) = p(x_t|z_{1:t})$. Therefore, the optimization process transforms into finding the $q(x_t)$ that minimizes the KLD~\eqref{eq::KLD}. In this work, we employ Gaussian VI (GVI), where $q(x_t)$ is chosen from a parametric family of Gaussian distributions~\cite{blei2017variational}. 
GVI deterministically finds the parameters of $q(x_t)$, leading to reduced computation time in high-dimensional problems compared to global approximation methods. 
 
Despite using a Gaussian distribution, the significance of GVI lies in its ability to select the best Gaussian fit for the approximation across the entire Gaussian parameter space, rather than relying on a local Gaussian approximation as in EKF and its variants. Consequently, GVI is expected to outperform EKF and similar filters. However, effectively solving \eqref{eq::KLD} presents a significant challenge, underscoring the contribution of this work. Specifically, the main contribution of this work is in designing an effective solver for \eqref{eq::KLD} by: (i) proposing a sequential GVI method that leverages sequential Bayesian principles within the GVI framework; (ii) demonstrating that the proposed method meets the natural expectation to yield the same solution as the information filter in linear models; and (iii) introducing a novel update equation for sequential GVI in nonlinear state estimation, using natural gradient descent~\cite{amari2016information}.  
We validated the performance of the proposed nonlinear state estimation approach in both simulations and real-world~experiments.

\emph{Related Work}: 
VI has been widely used in machine learning and reinforcement learning, e.g.,~\cite{rudner2022tractable,fellows2019virel}. Recently, VI has also been applied to state estimation with Bayesian filtering. For example, \cite{smidl2008variational} used recursive VI with sequential Monte Carlo, ignoring the state correlation. \cite{hu2018iterative} applied the proximal iterative method to minimize KLD, with the initial state estimated using the EKF. \cite{seo2023stein} employed particle-based VI to compute the MAP sequence estimation. Alternatively, \cite{barfoot2020exactly} applied GVI to batch state estimation, calculating the full posterior PDF on all states. \cite{courts2021gaussian} used GVI for nonlinear smoothing with a joint proposal distribution and also applied it to nonlinear filtering via marginalization. \cite{lambert2022continuous} applies GVI to both the prediction and correction steps of Kalman filtering. GVI has also been applied to other robotics applications such as motion planning~\cite{yu2023gaussian}, trajectory estimation~\cite{wong2020variational}, and localization~\cite{goudar2022gaussian}. In this study, we leverage sequential Bayesian principles within the GVI framework to optimize and track posterior PDFs over time without relying on traditional Bayesian filtering.

\section{Sequential Gaussian Variational Inference} \label{sec::SGVI}
In this work, we consider the problem of state estimation as finding $q(x_t) = \mathcal{N}(\mu_t, \Sigma_t)$\footnote{ $\mathcal{N}(\mu,\Sigma)$ is a Gaussian distribution with mean $\mu$ and covariance $\Sigma$.} that minimizes the KLD~\eqref{eq::KLD}. Specifically, we aim to find the optimal parameter $\theta_t := \{\mu_t, \Sigma_t^{-1}\}\in \Xi_{\theta_t} = \mathbb{R}^{n_x} \times \mathcal{S}_{++}^{n_x}$\footnotemark{}\footnotetext{$\mathcal{S}_{++}^{n_x}$ denotes the set of $n_x \times n_x$ symmetric positive definite matrices.} of the proposal distribution $q(x_t)$ to approximate the posterior PDF $p(x_t|z_{1:t})$. In what follows, for simplicity of notation, we write $q_{\theta_t}(x_t)$ simply as $q_{\theta_t}$.

Let $\mathcal{L}(\theta_t):=\text{KL}[q_{\theta_t}\|p(x_t|z_{1:t})]$. Then, substituting for $p(x_t|z_{1:t})$ in~\eqref{eq::KLD}, we arrive at
\begin{align*}
    &\mathcal{L}(\theta_t) \!=\!\! \!\int \!\! q_{\theta_t}\! \ln{\!\frac{q_{\theta_t} \int p(z_t|x_t)p(x_t|z_{1:t-1}) \text{d}x_t}{p(z_t|x_t) \!\int \!p(x_t|x_{t-1})p(x_{t-1}|z_{1:t-1}) \text{d}x_{t-1} }} \text{d}x_t. 
\end{align*}
To compute the optimal $\theta_t$, we propose a Sequential Gaussian Variational Inference (S-GVI) framework that uses $q_{\theta_{t\text{-}1}} = \mathcal{N}(\mu_{t-1}, \Sigma_{t-1})$ in place of $p(x_{t-1}|z_{1:t-1})$. It initializes with a Gaussian prior $p(x_0) = \mathcal{N}(\mu_0, \Sigma_0)$  for $q_{\theta_0}$, allowing us to write:
\begin{align}\label{eq::L_theta_0}
    \mathcal{L}(\theta_t) &= \int q_{\theta_t}  \ln{\frac{q_{\theta_t} p(z_t|z_{1:t-1})}{p(z_t|x_t) \int p(x_t|x_{t-1})q_{\theta_{t\text{-}1}} dx_{t-1} }} \text{d}x_t \nonumber \\
    &= \int q_{\theta_t} \ln{q_{\theta_t} }dx_t - \int q_{\theta_t} \ln{p(z_t|x_t)}\text{d}x_t\nonumber \\
    &~~~- \int q_{\theta_t}  \Bigl(\ln{ \int p(x_t|x_{t-1})q(x_{t-1}) dx_{t-1}}\Bigl) \text{d}x_t  \nonumber\\
    &~~~+ \ln{p(z_t|z_{1:t-1})}\nonumber\\
    &=\mathbb{E}_{q_{\theta_t}}[\ln q_{\theta_t}] - \mathbb{E}_{q_{\theta_t}}[\ln{p(z_t|x_t)}]- \nonumber\\ 
    &~~~~\mathbb{E}_{q_{\theta_t}}\!\bigr[\ln \mathbb{E}_{q_{\theta_{t\text{-}1}}}\! p(x_t|x_{t-1})]\bigr]+\ln{p(z_t|z_{1:t-1})},
\end{align}
where $\mathbb{E}[\cdot]$ denotes the expectation operator and the third term represents marginalization over $x_{t-1}$. The last term in \eqref{eq::L_theta_0} arises from the simplification $\int q_{\theta_t} \ln{p(z_t|z_{1:t-1})} \text{d}x_t = \ln{p(z_t|z_{1:t-1})}$, as $p(z_t|z_{1:t-1})$ is not dependent on $x_t$. Moreover, since $\ln{p(z_t|z_{1:t-1})}$ is independent of $\theta_t$, it can be omitted in the optimization problem defined to find $\theta^\star_t$:
\begin{align} 
\label{eq::Optimization_cost} 
    \theta_t^{\star} = \arg \! \min_{\!\!\!\!\! \theta_t \in \Xi_{\theta_t}} \mathcal{L}(\theta_t), 
\end{align}
where with an abuse of notation we redefine
\begin{equation}\label{eq::Optimization_cost_final}
\begin{split}
   \mathcal{L}(\theta_t) = &\mathbb{E}_{q_{\theta_t}}[\ln q_{\theta_t}] - \mathbb{E}_{q_{\theta_t}}\![\ln{p(z_t|x_t)}]  \\
   &\qquad\qquad~-\mathbb{E}_{q_{\theta_t}}\!\bigr[\ln \mathbb{E}_{q_{\theta_{t\text{-}1}}}\![p(x_t|x_{t-1})]\bigr].
\end{split}
\end{equation}
Notably, the first term in~\eqref{eq::Optimization_cost_final}, by direct integration, equates~to 
\begin{align}
      \mathbb{E}_{q_{\theta_t}}[\ln q_{\theta_t}] 
      &= -\frac{1}{2} \ln\bigl( (2\pi)^{n_x}|\Sigma_t| \bigl)\,-\,\frac{n_x}{2}.
\end{align}
For nonlinear motion and measurement models, computing the remaining two terms in~\eqref{eq::Optimization_cost_final} is more complex, as addressed in Section~\ref{sec::SGVI_Nonlinear}. For linear systems, though, these terms can be computed in closed form (see Appendix B), enabling an analytic solution for \eqref{eq::Optimization_cost_final}. This solution, derived in Appendix A, aligns exactly with the optimal information filter for linear systems. Such promising results are typically expected from the approximate state estimation methods proposed for nonlinear systems.

\begin{rem}
    The main differences from the existing work are as follows. In~\cite{courts2021gaussian}, $q(x_t, x_{t-1})$ is derived from $p(x_t, z_t | x_{t-1})$, taking into account the uncertainty in the posterior PDF of $x_{t-1}$. Then $q(x_t)$ is obtained by marginalizing $q(x_t, x_{t-1})$ over $x_{t-1}$. The update equation for the optimization process is solved using standard optimization techniques that employ exact first and second derivatives. In~\cite{barfoot2020exactly}, $q(x_t)$ is derived from the entire trajectory using a batch optimization process. Assuming a factorized joint likelihood of the state and the observed data, $q(x_t)$ is marginalized out from the batch solution, resulting in an efficient optimization scheme. \cite{lambert2022continuous} applies GVI separately to the prediction step, $p(x_t, x_{t-1})$, and the correction step, $p(x_t | z_t)$, to handle continuous-time dynamical processes and observed data in discrete time.\boxend 
\end{rem}

\section{S-GVI for Nonlinear State Estimation}
\label{sec::SGVI_Nonlinear}
In this section, we introduce a novel update equation for $\theta_t$, derived from minimizing \eqref{eq::Optimization_cost_final} w.r.t. $\theta_t$ for nonlinear systems described by\footnotemark{}\footnotetext{The models correspond to the form $x_t=f_{t|t-1}(x_{t-1})+v_{t-1}$, where $v_{t-1}\sim\mathcal{N}(0, Q_{t-1})$, and $z_t=h_t(x_t)+r_t$, where $r_t\sim\mathcal{N}(0, R_t)$~\cite{sarkka2023bayesian}.}:
\begin{align}
\label{eq::NSSM}
    x_t \sim \mathcal{N}(f_{t|t-1}(x_{t-1}),\, Q_{t-1}),~~z_t \sim \mathcal{N}(h_t(x_t),\, R_t), 
\end{align}
where $f_{t|t-1}(x_{t-1})$ and $h_t(x_t)$ represent the nonlinear model in the state transition and likelihood function, respectively. The choice of a nonlinear model with additive Gaussian noise is common in robotic applications~\cite{dellaert2021factor, barfoot2024state}. For the observed data $z_t$, we assume that $z_t$ follows conditional independence given $x_t$ at time $t$. In what follows, to simplify the notation, we omit $t$ and $t|t-1$ in the system parameters $(f_{t|t-1}, h_t, Q_{t-1}, R_t)$.

To solve for $\theta^\star_t$ in~\eqref{eq::Optimization_cost_final}, three essential steps are required: (i) a statistical approximation for the marginalization term $\mathbb{E}_{q_{\theta_{t\text{-}1}}}[p(x_t|x_{t-1})]$, (ii) a deterministic numerical method to efficiently approximate expectation terms, and (iii) an efficient numerical solver to solve for $\theta^\star_t$.

\subsection{Statistical Approximation for Marginalization}
To compute $\mathbb{E}_{q_{\theta_{t\text{-}1}}}[p(x_t|x_{t-1})]$, we propose using \emph{Statistical Linear Regression} (SLR) theory~\cite{arasaratnam2007discrete}, which offers a flexible and general framework for approximation. SLR accounts for the uncertainty in the random variable during linearization, making the linearized function statistically more accurate than the Taylor expansion about a single point, such as the mean (see details in Chapters 9.2 and 9.4 of~\cite{sarkka2023bayesian}). 

SLR approximates the nonlinear state transition model in~\eqref{eq::NSSM} as $f(x_{t-1}) \approx Fx_{t-1}+b+e$ where $e \sim \mathcal{N}(0, \Lambda)$. By accounting for the linearization error, which arises from the nonlinearity of $f(x_{t-1})$, within $\Lambda$, the SLR technique is shown to provide a consistent estimate, see details in~\cite{arasaratnam2007discrete}. Given  $q_{\theta_{t\text{-}1}} = \mathcal{N}(\mu_{t-1}, \Sigma_{t-1})$ from the previous time step $t-1$, the unknown parameters $\{F, b\}$ are chosen to minimize the mean square error as
\begin{align}\label{eq::MSE_FB}
      \!\!\min_{F,b} \mathbb{E}\bigl[(f(x_{t-1})\!-\!Fx_{t-1}\!-\!b)\!^{\top}\!(f(x_{t-1})\!-\!Fx_{t-1}\!-\!b)\bigl].\!\!\!\!
\end{align}
Then, the error covariance $\Lambda$ is computed from 
\begin{align}\label{eq::Lambda}
      \Lambda \!=\! \mathbb{E}\bigl[(f(x_{t-1})\!-\!Fx_{t-1}\!-\!b)(f(x_{t-1})\!-\!Fx_{t-1}\!-\!b)\!^{\top}\bigl].\!\!
\end{align}
 Using the first-order optimality condition for~\eqref{eq::MSE_FB}, the closed-form solution of~\eqref{eq::MSE_FB} and~\eqref{eq::Lambda} are obtained~\cite{arasaratnam2007discrete}:
\begin{align}
      &F = C^{\top}\Sigma_{t-1}^{-1},~b = \mu_R \!-\! F\mu_{t-1},~ \Lambda = S - F\Sigma_{t-1}F^{\top}\!\!\!,\!
\end{align}
where
\begin{subequations}\label{eq:prior_para}
\begin{align} 
    C&=\mathbb{E}_{q_{\theta_{t\text{-}1}}}[(x_{t-1}-\mu_{t-1})(f(x_{t-1})-\mu_R)], \label{eq:prior_para1}\\
    \mu_R &= \mathbb{E}_{q_{\theta_{t\text{-}1}}}[f(x_{t-1})], \label{eq:prior_para2}\\
    S&=\mathbb{E}_{q_{\theta_{t\text{-}1}}}[(f(x_{t-1})-\mu_R)(f(x_{t-1})-\mu_R)\!^{\top}] \!+\! Q. \label{eq:prior_para3}
\end{align}
\end{subequations}
Thus, $p(x_t|x_{t-1})\approx\mathcal{N}(x_t|Fx_{t-1}+b,\Lambda)$. Consequently, we can compute the marginalization term analytically as follows:
\begin{align}
    &\mathbb{E}_{q_{\theta_{t\text{-}1}}}[p(x_t|x_{t-1})] \nonumber \\ 
    &\approx \int \mathcal{N}(x_t|Fx_{t-1}+b,\, \Lambda)~\mathcal{N}(x_{t-1}|\mu_{t-1},\,\Sigma_{t-1}) dx_{t-1} \nonumber \\
    &= \int \mathcal{N}\biggl(\!
        \begin{bmatrix}
        x_{t-1} \\
        x_t
        \end{bmatrix}
        | 
        \begin{bmatrix}
        \mu_{t-1} \\
        \mu_R
        \end{bmatrix}
        ,
        \begin{bmatrix}
        \Sigma_{t-1} & C \\
        C^{\top} & S
        \end{bmatrix}
        \biggl) dx_{t-1} \nonumber \\
    &= \mathcal{N}(x_t\,|\,\mu_R, S).
\end{align}
Note that to compute $S$, the state transition function $f$ does not need to be differentiable. This statistical approximation for the marginalization term $\mathbb{E}_{q_{\theta_{t\text{-}1}}}[p(x_t|x_{t-1})]$ in~\eqref{eq::Optimization_cost_final} enables efficient gradient computation in the optimization procedure described next.

\subsection{Numerical solver}
To solve \eqref{eq::Optimization_cost}, we propose using the \emph{natural gradient descent} (NGD) approach. NGD is typically employed for optimizing probabilistic models and involves the use of the ``natural gradient," which is defined as the gradient multiplied by the inverse of the Fisher Information Matrix (FIM) of the model. Unlike the fixed structure of Euclidean space, the parameter space of a Gaussian distribution is a Riemannian manifold, with its geometry dynamically changing based on the values of the parameters $\theta_t$. This geometric structure is characterized by the FIM, which serves as the metric tensor of the manifold. Consequently, the distance between parameter points varies according to the FIM, necessitating consideration of this geometry when updating the parameters. As a result, compared to standard gradient descent in Euclidean space, NGD often requires fewer iterations, making it a more efficient and attractive alternative. For more information on NGD, see \cite{amari2016information,khan2017variational}\footnotemark{}\footnotetext{\cite{khan2017variational} demonstrated the NGD update equations are equivalent to the mirror descent step in the expectation parameter space $\{\mathbb{E}[x], \mathbb{E}[x^2]\}$ without directly computing the inverse of FIM.}.

To use NGD for solving optimization problem~\eqref{eq::Optimization_cost}, the descent direction for $\Sigma_t^{-1}$ and $\mu_t$ are obtained by gradients multiplied by the inverse of the FIM  given respectively by 
$F_{\Sigma_t^{-1}}^{-1}\nabla_{\Sigma_t^{-1}}\mathcal{L}(\theta_t)=-2\,\nabla_{\Sigma_t}\mathcal{L}(\theta_t)$ and $F_{\mu_t}^{-1}\nabla_{\mu_t}\mathcal{L}(\theta_t)=\Sigma_t\nabla_{\mu_t}\mathcal{L}(\theta_t)$. Using these descent directions with step size $\beta_t\in\mathbb{R}_{>0}$ the iterative NGD sequential update equations are \begin{subequations}\label{eq::update_mirror}
\begin{align}
      \Sigma_t^{-1} &\leftarrow \Sigma_t^{-1} + 2\beta_t \nabla_{\Sigma_t}\mathcal{L}(\theta_t), \label{eq::mirror1} \\
      \mu_t &\leftarrow \mu_t - \beta_t\Sigma_t\nabla_{\mu_t}\mathcal{L}(\theta_t). \label{eq::mirror2}
\end{align}
\end{subequations}


Using Bonnet and Price's theorems~\cite{opper2009variational}, the gradient terms for the expectation w.r.t. $\theta_t$ are converted into the expectation terms for the gradient and Hessian w.r.t. $x_t$:
\begin{subequations}
\begin{align}
      &\nabla_{\mu_t}\mathcal{L}(\theta_t)=\nabla_{\mu_t}\mathbb{E}_{q_{\theta_t}}[l(\theta_t)] = \mathbb{E}_{q_{\theta_t}}[\nabla_{x_t}l(\theta_t)], \\
      &\nabla_{\Sigma_t}\mathcal{L}(\theta_t)=\nabla_{\Sigma_t}\mathbb{E}_{q_{\theta_t}}[l(\theta_t)] = \frac{1}{2}\mathbb{E}_{q_{\theta_t}}[\nabla_{x_tx_t}l(\theta_t)],
\end{align}
\end{subequations}
where $l(\theta_t):=\ln q_{\theta_t} - \ln{p(z_t|x_t)} - \ln \mathbb{E}_{q_{\theta_{t\text{-}1}}}[p(x_t|x_{t-1})]$. Next, assuming that the measurement model function $h$ is at least first-order differentiable, we can write 
\begin{subequations} \label{eq::gradient}
\begin{align}
    &\nabla_{x_t}l(\theta_t) = -H_J^{\top}R^{-1}(z_t-h(x_t))-S^{-1}(\mu_R-x_t) \nonumber \\
    &\qquad\qquad~~ - \Sigma_{t}^{-1}(x_t-\mu_t), \\
    &\nabla_{x_tx_t}l(\theta_t) = H_J^{\top}R^{-1}H_J + S^{-1} - \Sigma_{t}^{-1},
\end{align}
\end{subequations}where $H_J = \nabla_{x_t} h(x_t)$ is the Jacobian matrix of $h$, and $S$ and $\mu_R$ are given in~\eqref{eq:prior_para}. Subsequently, the update equations~\eqref{eq::update_mirror} become 
\begin{subequations}\label{eq:proposed_update}
\begin{align}
    &\Sigma_t^{-1} \leftarrow (1-\beta_t)\Sigma_t^{-1} \!+\! \beta_t\bigl(\mathbb{E}_{q_{\theta_t}}[H_J^{\top}R^{-1}H_J] \!+ S^{-1}\bigl),\!\! \label{eq:proposed_update1} \\
    &\mu_t \leftarrow \mu_t + \beta_t\,\Sigma_t\times \label{eq:proposed_update2} \nonumber\\
    &\qquad\Bigl(\mathbb{E}_{q_{\theta_t}}\bigl[H_J^{\top}R^{-1}\bigl(z_t-h(x_t)\bigl)\bigl] + S^{-1}(\mu_R-\mu_t)\Bigl),\!\! 
\end{align}
\end{subequations}
where the last term in~\eqref{eq:proposed_update2} is derived from $\mathbb{E}_{q_{\theta_t}}\bigl[S^{-1}(\mu_R-x_t)] = S^{-1}(\mu_R-\mu_t)$ as $S^{-1}$ and $\mu_R$ are independent of $x_t$. Algorithm~\ref{alg:SGVI} summarizes our proposed S-GVI update step to calculate $\theta^\star_t$.

The update equations \eqref{eq:proposed_update} involve two expectation terms $\mathbb{E}_{q_{\theta_t}}[\cdot]$ and $\mathbb{E}_{q_{\theta_{t\text{-}1}}}[\cdot]$ that cannot be computed analytically, which require approximation techniques. One simple method for approximation is the first-order delta method, which evaluates the expectation at the mean, though this sacrifices accuracy~\cite{ver2012invented}. Alternatively, Monte Carlo (MC) sampling can be used, but it is computationally inefficient due to its dependence on random sampling. In our work, we apply the \emph{quadrature rules}\footnotemark{}\footnotetext{There are various quadrature rules in the field of Bayesian filtering, such as Gauss-Hermite quadrature~\cite{ito2000gaussian} and the unscented transform~\cite{julier2004unscented}.} using sigma points, a deterministic sampling approach, to computationally efficiently approximate the expectation of functions,  i.e., $\mathbb{E}_{q_{\theta}}[g(x)]\approx \sum\nolimits_{i=1}^m w^i g(x^i),$ where $w^i \in \real$ is the weight satisfying $\sum\nolimits_{i=1}^m w^i= 1$, $x^i=\mu+\sqrt{\Sigma}\,\zeta^i$ are sigma-points, $\zeta^i$ are unit sigma-points and $m$ is the total number of sigma-points (see details in~\cite{sarkka2023bayesian}).

We close this section with some key insights on the update equations in~\eqref{eq:proposed_update}:
\begin{enumerate}
    \item In~\eqref{eq:proposed_update1}, for $0 < \beta_t \leq 1$, $\Sigma_t^{-1}$ is updated by a convex combination of the previous $\Sigma_t^{-1}$ and the gradient, similar to the gradient momentum method in stochastic optimization. The precision $\Sigma_t^{-1}$ increases at each iteration by the precision contributions of the measurement model, $\mathbb{E}_{q_{\theta_t}}[H_J^{\top}R^{-1}H_J]$, and the state transition model, $S^{-1}$, where $S=\mathbb{E}_{q_{\theta_{t\text{-}1}}}[(f(x_{t-1})-\mu_R)(f(x_{t-1})-\mu_R)^{\top}] + Q$. Moreover, when $0 < \beta_t \leq 1$, and given that $Q > 0$ and $\Sigma_0 > 0$, we are guaranteed that $\Sigma_t^{-1} > 0$.    
    \item The gradient update~\eqref{eq:proposed_update2} consists of summing the weighted average deviations between the data $z_t$ and the previous mean $\mu_R$. It is important to note that the parameters are updated w.r.t. the posterior PDF space. Consequently, $H_J$, $\mathbb{E}_{q_{\theta_t}}[\cdot]$, and $\mu_t$ are computed considering both the motion model and the data $z_t$. This approach differs from approximate Bayesian filtering, which typically involves prediction and correction steps; for example, the prior PDF is computed during the prediction step, and then $H_J$ is computed w.r.t. this prior PDF, without considering the data $z_t$.
    Lastly, the update is \emph{geometrically scaled} by the updated $\Sigma_t$ (i.e., FIM) as an inner product, which takes into account the geometry of the Gaussian parameters.
    \item In~\eqref{eq:proposed_update}, the iterative procedure of updating $\mu_t$ and $\Sigma_t^{-1}$, unlike the solution in linear models~\eqref{eq::update_liner}, involves previously updated parameters indirectly influencing subsequent updates via $H_J$ and $\mathbb{E}_{q_{\theta_t}}[\cdot]$. Additionally, the update equations involve averaging (i.e., the expectation $\mathbb{E}_{q_{\theta_t}}[\cdot]$) of the iterated values of $x_t$, rather than using a single point. Gradient averaging improves the convergence rate compared to not using averaging~\cite{schmidt2017minimizing,rostami2024forward}.
\end{enumerate}

\begin{algorithm}[t]
\caption{The update step of S-GVI}
\label{alg:SGVI}
\SetAlgoLined
\LinesNumbered
\DontPrintSemicolon
\SetKwInOut{Input}{Input}\SetKwInOut{Output}{Output}
\Input{$\theta_{t-1}^{\star} = \{\mu_{t-1},\Sigma_{t-1}^{-1}\}$}
\Output{$\theta_t^{\star} = \{\mu_t,\Sigma_t^{-1}\}$}
\BlankLine
Initialize $\theta_t = \theta_{t-1}^{\star}$ \\
{$\mu_R, S \stackrel{\eqref{eq:prior_para}}{\longleftarrow} \{x_{t-1}^i, w_{t-1}^i\}_{i=1}^m \longleftarrow \theta_{t-1}^{\star}$} \\
\algorithmicwhile{~($\bar{e}>\epsilon)$\footnotemark{}} \textbf{do}\\
{$\qquad\Sigma_t^{-1} \longleftarrow$ \eqref{eq:proposed_update1}} \\
{$\qquad\mu_t \longleftarrow$\eqref{eq:proposed_update2} given $\Sigma_t^{-1}$} \\
\algorithmicend \\
\end{algorithm}
\footnotetext{$\bar{e}$ and $\epsilon$ are the iteration termination conditions. We use the relative amount of change, i.e., $\bar{e}:=\Bigl|\frac{\theta^{k+1}-\theta^k}{\theta^k}\Bigl|$ where $k$ is the iteration step. For the matrix case, we can use an appropriate matrix norm, such as the Frobenius norm, infinity norm, or $2$-norm~\cite{horn2012matrix}.}

\section{Experiments}\label{sec::experiment}

We evaluated the proposed S-GVI against maximum a posteriori (MAP) estimation using the iterated Extended Kalman Filter (iEKF), wherein a first-order Taylor expansion is applied iteratively to find (local) MAP solution in the filtering process\footnotemark{}.\footnotetext{The iterative procedure can be interpreted as the Gauss-Newton method for finding the MAP estimate at the correction step~\cite{sarkka2023bayesian}.} For hyper-parameters, we set the constant step size $\beta_t=1$, the iteration termination condition $\epsilon=0.02$. For the stopping criterion of the multi-dimensional covariance matrix, we used the Frobenius norm. Furthermore, we applied the unscented transform as the quadrature rule, with the number of sigma points given by $m = 2 n_x + 1$ and the parameters $\kappa = 3 - n_x$, $\alpha = 1$, and $\beta^s = 0.1$ in both simulations and real-world experiments (see details in~\cite{julier2004unscented}). The performance of the two approaches is compared using the root-mean-square error (RMSE) metric over the trajectory. For real-world experiments, the error with $3$-sigma bounds and the normalized estimation error squared (NEES) metric are used additionally.

\subsection{Simulation: Nonlinear One-dimensional System}
First, we consider a benchmark problem from~\cite{garcia2016iterated}, where the system model~\eqref{eq::NSSM} is given by
\begin{subequations}
\begin{align}
    x_t &= 0.9 x_{t-1} \!+\! \frac{10x_{t-1}}{1+x_{t-1}^2} \!+ 8\cos{\bigl(1.2(t\!-\!1)\bigl)} + v_{t-1},\! \label{eq::sim_m1} \\ \label{eq::sim_m2} 
    z_t &= 0.05 x_t^3 + r_t, 
\end{align}
\end{subequations}
with $v_{t-1}\sim\mathcal{N}(0, Q)$, $r_t\sim\mathcal{N}(0, R)$ and $p(x_0) = \mathcal{N}(5, 2^2)$. The significance of this example is that the term $x_t^3$ in~\eqref{eq::sim_m2} introduces a higher degree of nonlinearity due to its faster growth rate, broader behavioral spectrum, and rapid error propagation.

\begin{figure}[!t]
    \centering
    \begin{minipage}[t]{0.225\textwidth}
        \centering
        \includegraphics[width=\textwidth]{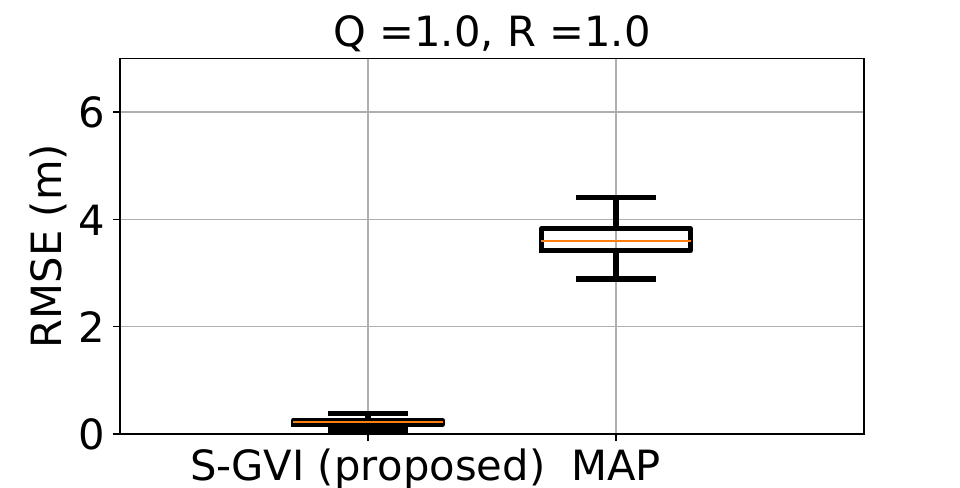}
        {{\scriptsize S-GVI (0.270m), MAP (3.664m)}}
    \end{minipage}
    \hskip\baselineskip
    \begin{minipage}[t]{0.225\textwidth}
        \centering
        \includegraphics[width=\textwidth]{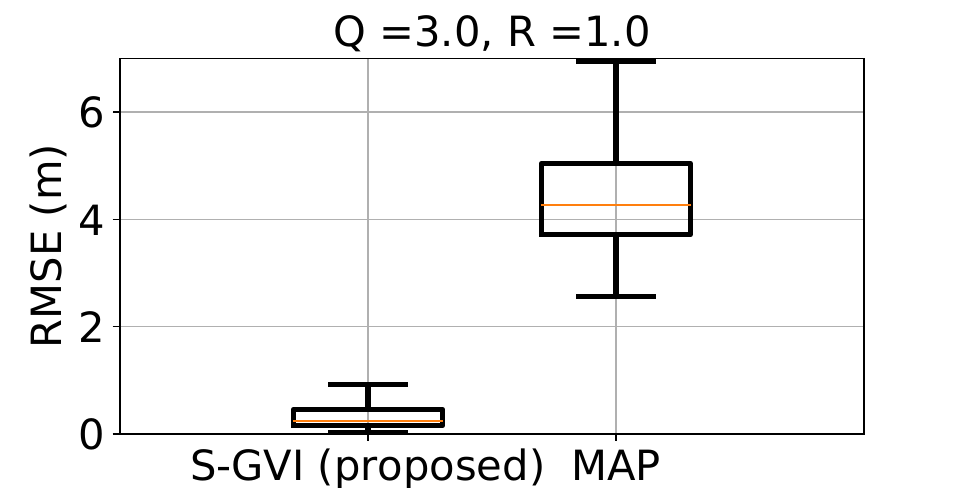}
        {{\scriptsize S-GVI (0.353m), MAP (4.712m)}}
    \end{minipage}
    \hskip\baselineskip
    \begin{minipage}[t]{0.225\textwidth}
        \centering
        \includegraphics[width=\textwidth]{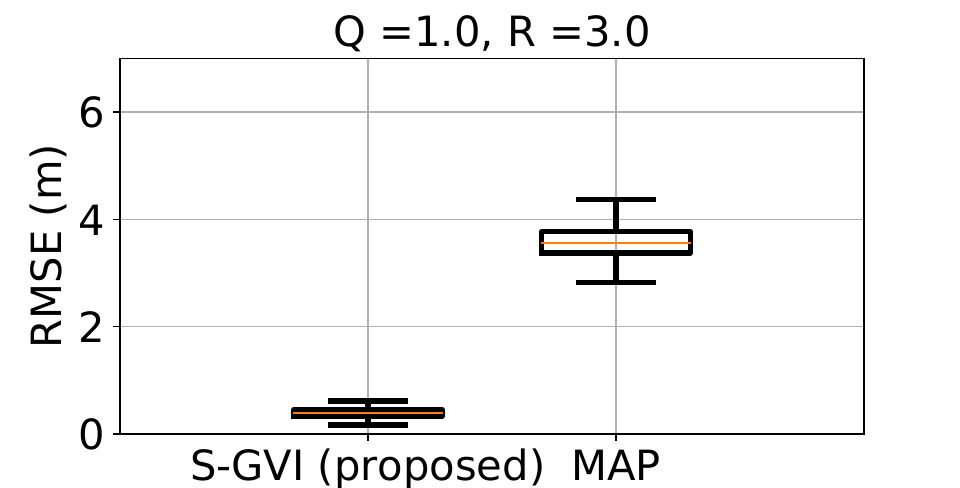}
        {{\scriptsize S-GVI (0.417m), MAP (3.592m)}}
    \end{minipage}
    \hskip\baselineskip
    \begin{minipage}[t]{0.225\textwidth}
        \centering
        \includegraphics[width=\textwidth]{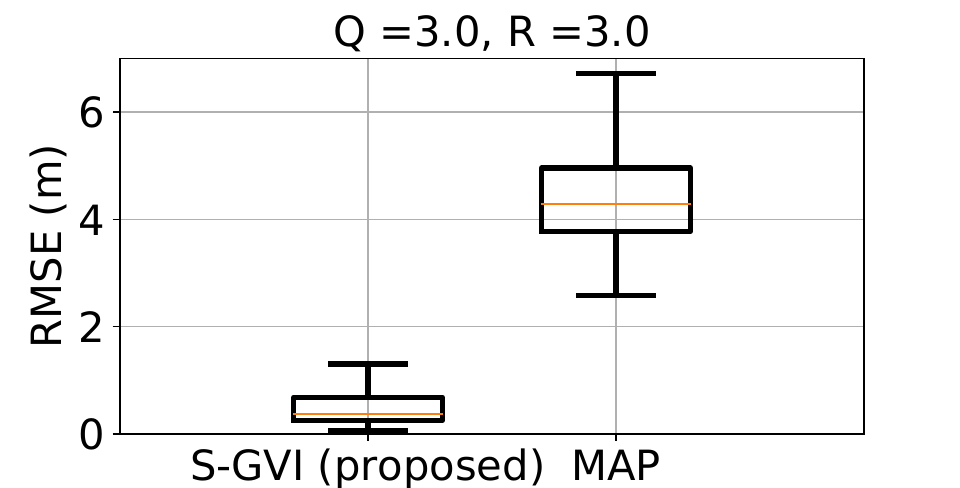}
        {{\scriptsize S-GVI (0.486m), MAP (4.381m)}}
    \end{minipage}
    \caption{{\small RMSE of the proposed S-GVI and MAP estimation over all Monte Carlo runs for different sets of $\{Q, R\}$. The mean values are provided in ($\cdot$). The S-GVI significantly outperforms the MAP estimation in all simulation sets, indicating that S-GVI is robust against high nonlinearity.}}
    \label{fig:Exp_Result1}
\end{figure}
To evaluate the algorithms, we generated $1,000$ independent trajectories, each with $50$ time steps. We calculated the RMSE of the state estimates averaged over all Monte Carlo runs for both S-GVI and MAP estimation. The RMSE was compared across different sets of $\{Q, R\}$. The results in Fig.~\ref{fig:Exp_Result1} demonstrate that S-GVI significantly outperforms MAP estimation, indicating that S-GVI is robust against high nonlinearity. This superiority of S-GVI can be attributed to its optimization in the posterior space without separating the steps for mean and covariance updates, unlike MAP, which requires iterative update steps based on the predicted mean and covariance from the propagation step. However, it is important to note that the one-dimensional case does not benefit from geometric scaling as much as the high-dimensional case (see results in Section~\ref{sec::real_exp}), where each axis is scaled by the FIM as an inner product. Consequently, the average number of iterations for MAP ($2.939$) is lower than for S-GVI ($6.964$).

\subsection{Real-world Experiment: Range-only Localization}
\label{sec::real_exp}

\begin{figure}[t]
    \centering
    \includegraphics[width=0.42\textwidth]{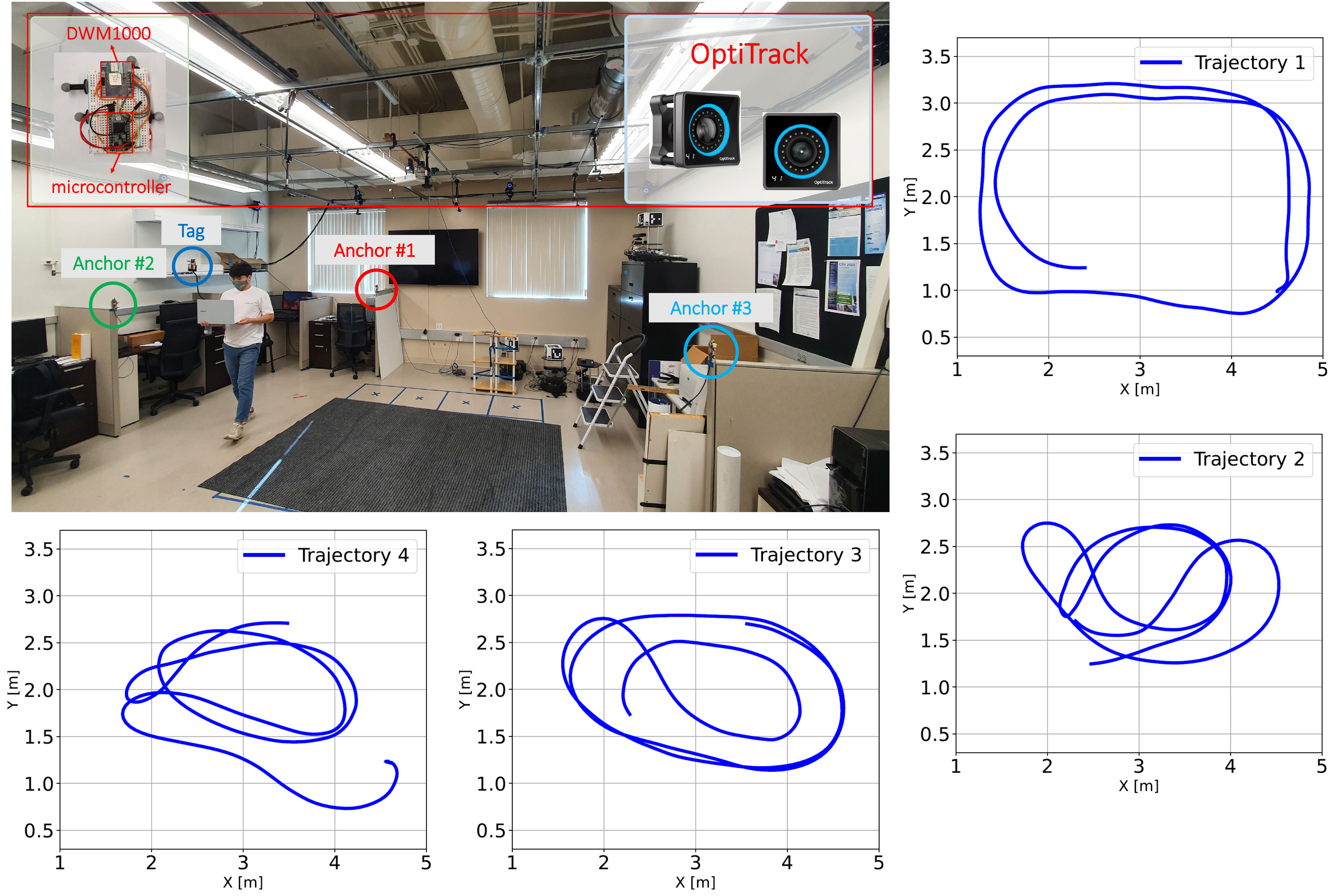}
    \caption{{\small Experimental setup and $4$ experiments with $300$ time steps.}}
    \label{fig:Exp}
\end{figure}

In this experiment, a human agent equipped with a DWM1000 ultra-wideband (UWB) transceiver walked in an indoor environment with three UWB anchors installed, as shown in Fig.~\ref{fig:Exp}. We performed $4$ experiments, each with $300$ time steps. The UWB system provided range measurements at a frequency of $10$ Hz. For comparison with a reference trajectory, we used an OptiTrack optical motion capture system, which uses $12$ infrared cameras to achieve a high-precision trajectory with an accuracy of $10^{-4}$m and a sampling rate of $120$ Hz.

We aim to localize a target without proprioceptive sensors such as IMUs or encoders. Thus, we used a discrete-time version of the coordinated turn model with polar velocity, as established in the tracking field~\cite{gustafsson1996best}: \begin{align}
\label{eq::Trans_Prob}
    \begin{bmatrix}
        x_t \\
        y_t \\
        v_t \\
        h_t \\
        w_t
    \end{bmatrix}
    \!\!\!=\!\!\! 
    \begin{bmatrix}
    x_{t-1} \!+\! \frac{2v_{t-1}}{w_{t-1}}\sin{(\frac{w_{t-1}T}{2})} 
    \cos{(h_{t-1}+\frac{w_{t-1}T}{2})} \\
    y_{t-1} \!+\! \frac{2v_{t-1}}{w_{t-1}}\sin{(\frac{w_{t-1}T}{2})} 
    \sin{(h_{t-1}+\frac{w_{t-1}T}{2})} \\
    v_{t-1} \\
    h_{t-1} + w_{t-1}T \\
    w_{t-1} \!\!
    \end{bmatrix}\!\!\!,\!\!\!
\end{align} 
\begin{align}
\label{eq::Trans_Process}
    &Q_{t-1}= \\
    &~~~
    \begin{bmatrix}
    \frac{T^2}{2}\cos(h_{t-1})~0 \\
    \frac{T^2}{2}\sin(h_{t-1})~0 \\
    ~~~~~T\qquad~~~0 \\
    ~~~~~~~0\qquad~~~\frac{T^2}{2} \\
    ~~~~~~0\qquad~~~~T \!\!
    \end{bmatrix}\!\!\!
    \begin{bmatrix}
    Q_a~~0 \\
    ~~0~~~Q_{\alpha}
    \end{bmatrix}\!\!\!
    \begin{bmatrix}
    \frac{T^2}{2}\cos(h_{t-1})~0 \\
    \frac{T^2}{2}\sin(h_{t-1})~0 \\
    ~~~~~T\qquad~~~0 \\
    ~~~~~~~0\qquad~~~\frac{T^2}{2} \\
    ~~~~~~0\qquad~~~~T \!\!
    \end{bmatrix}^{\top}\!\!\!\!\!\!,\!\!\!\nonumber
\end{align}
where the state vector includes the 2D position $(x, y)$ [m], the velocity magnitude $v$ [m/s], the heading angle $h$ [rad], and the turn rate $w$ [rad/s]. The discrete-time sampling period is denoted by $T$. In the process noise covariance model~\eqref{eq::Trans_Process}, $Q_a$ and $Q_{\alpha}$ are the linear and rotational acceleration noise, respectively. In our experiment, we set $Q_a = 2.0^2$ and $Q_{\alpha} = 0.01^2$. The initial condition for each experiment is provided by OptiTrack, and the initial covariance, $\Sigma_0=\text{diag}([0.05^2, 0.05^2, 0.01^2, 0.01^2, 0.01^2])$, is the same for all experiments.

The likelihood function of line-of-sight range measurements is given by:
\begin{align}
\label{eq::Obs_Prob1} 
    z_{t,l} = \|\Gamma_l\,-\,P_t\|_{2} + r_t,
\end{align}
where $z_{t,l} \in \mathbb{R}_{\geq 0}$ represents the measurement value from the $l$-th anchor $\Gamma_l$, the 2D position vector, at time index $t$. Here, $P_t = [x_t, y_t]^{\top}$ is the 2D position vector, $\|\cdot\|_2$ denotes the $l_2$ norm, and $r_t \sim \mathcal{N}(0, 0.5^2)$.

\begin{figure}[!t]
    \centering
    \begin{minipage}[t]{0.22\textwidth}
        \centering
        \includegraphics[width=\textwidth]{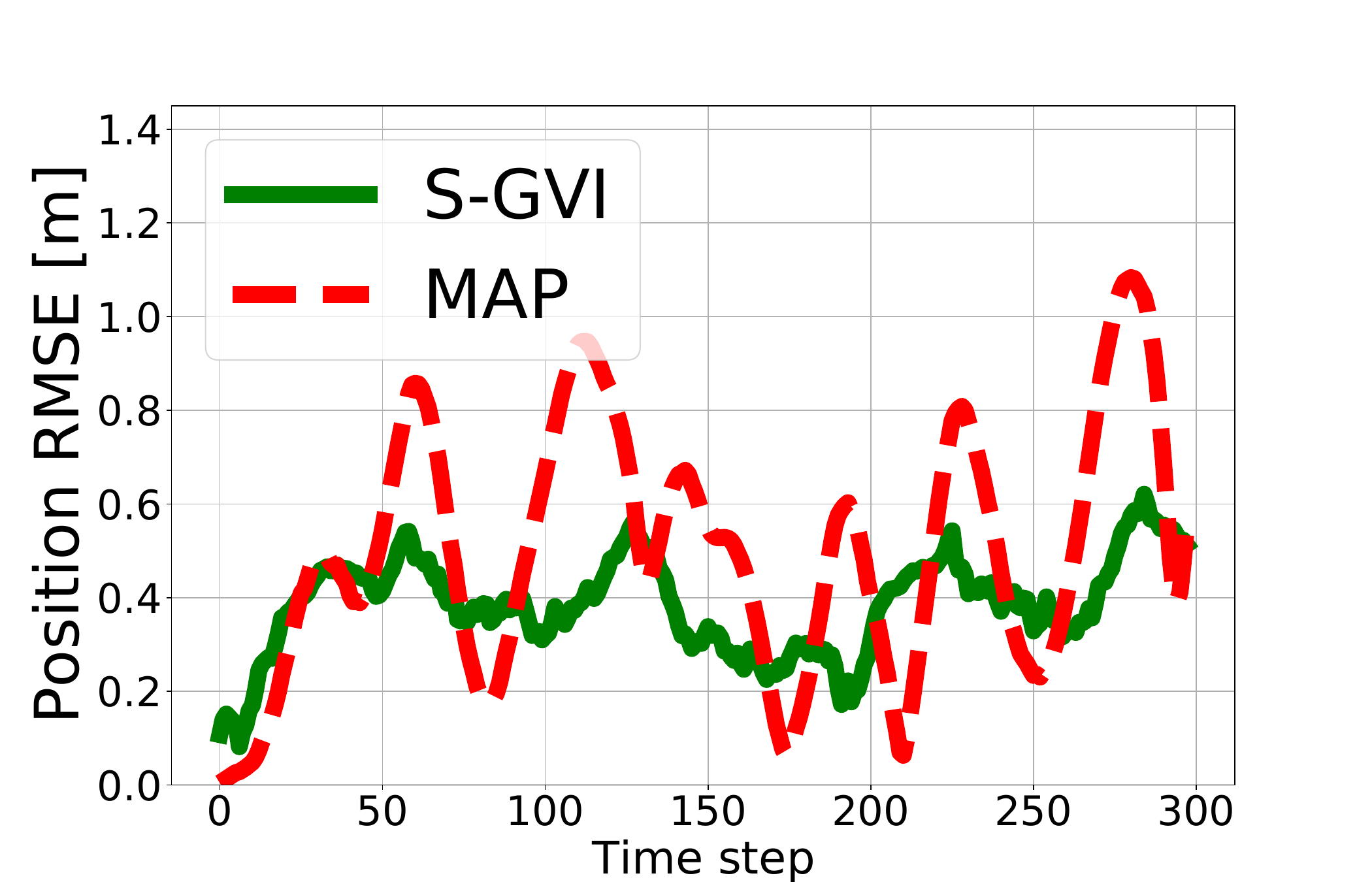}
        {{\scriptsize (a) Experiment 1}}
    \end{minipage}
    \hskip\baselineskip
    \begin{minipage}[t]{0.22\textwidth}
        \centering
        \includegraphics[width=\textwidth]{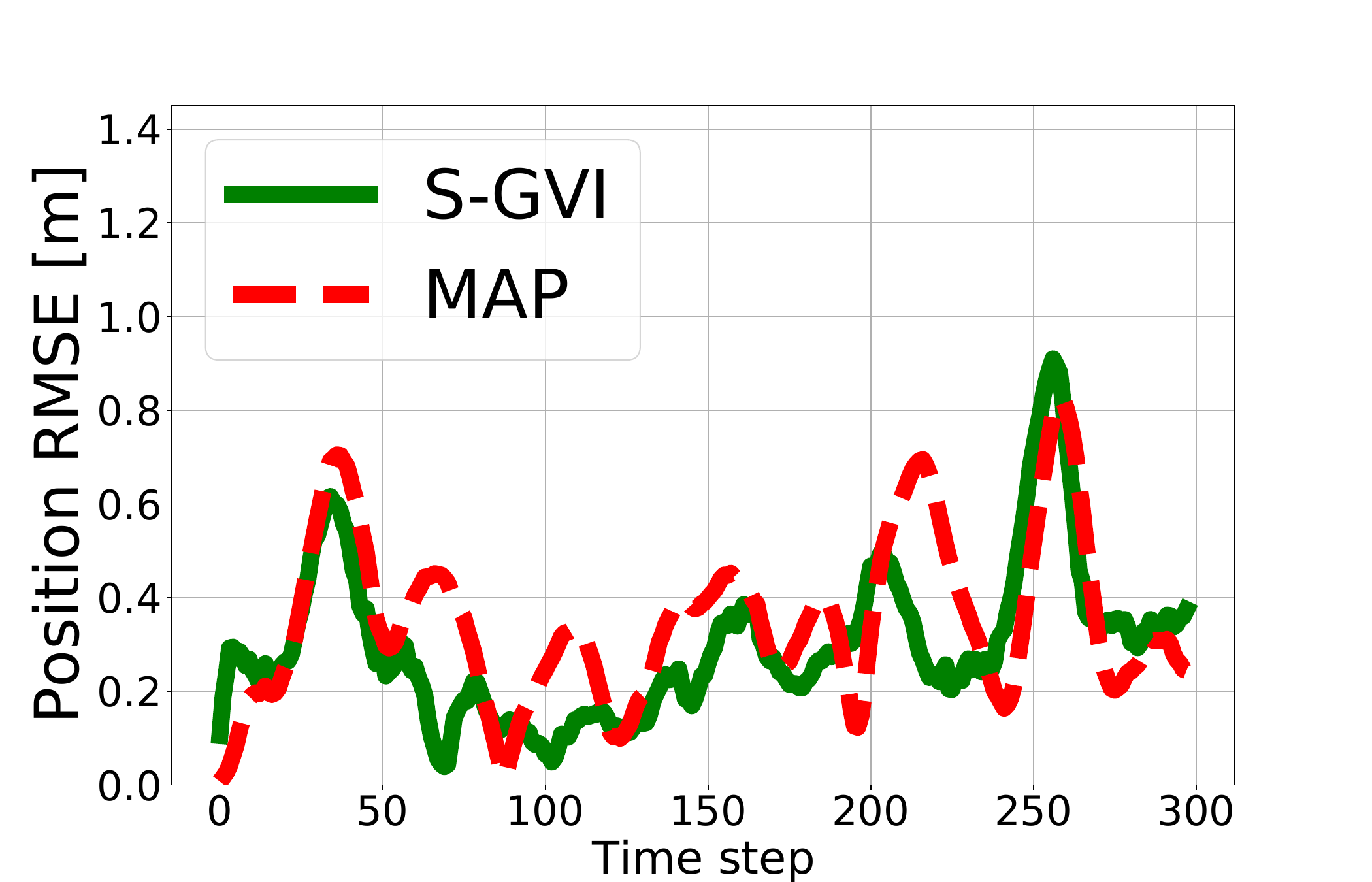}
        {{\scriptsize (b) Experiment 2}}
    \end{minipage}
    \hskip\baselineskip
    \begin{minipage}[t]{0.22\textwidth}
        \centering
        \includegraphics[width=\textwidth]{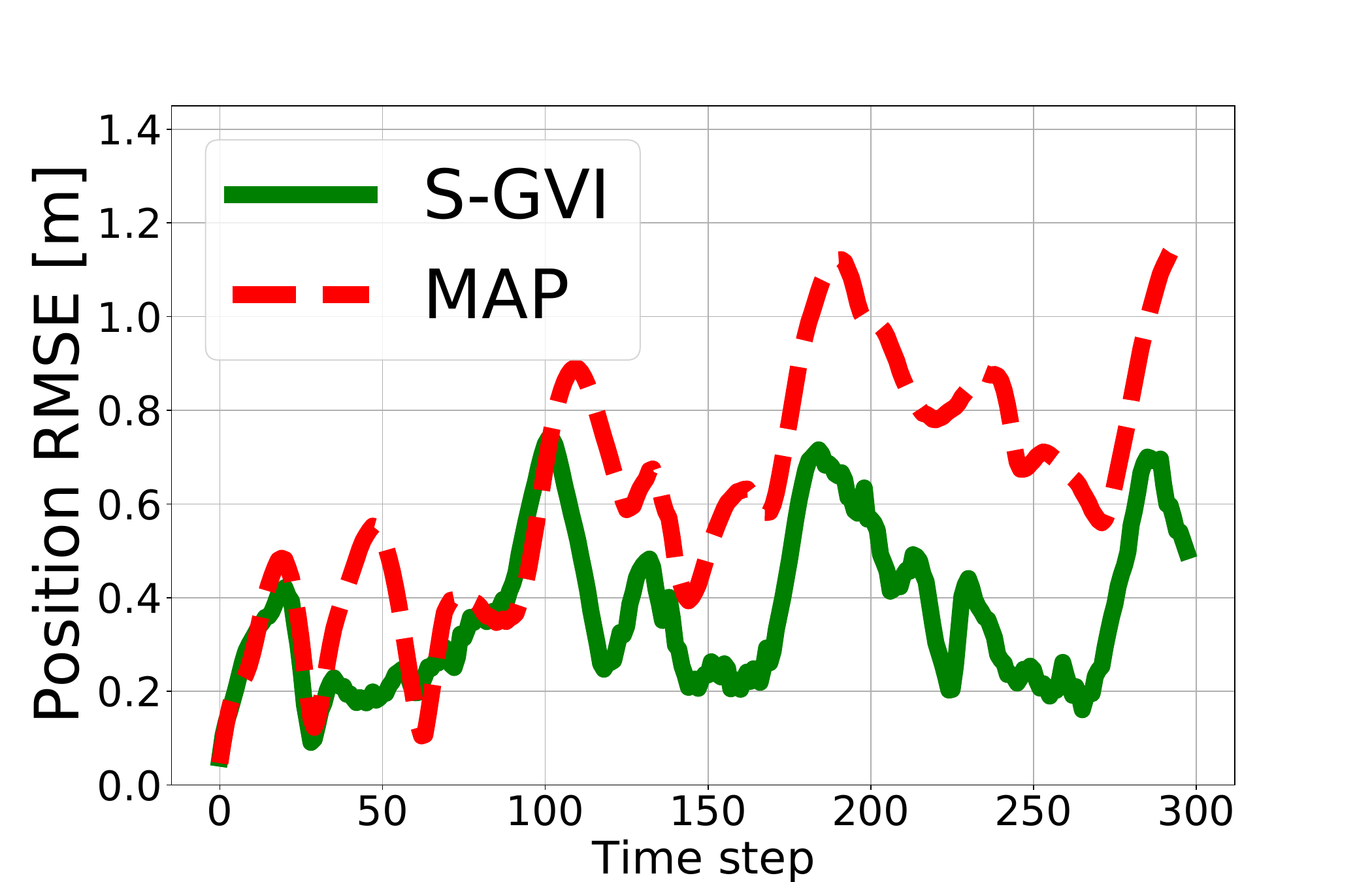}
        {{\scriptsize (c) Experiment 3}}
    \end{minipage}
    \hskip\baselineskip
    \begin{minipage}[t]{0.22\textwidth}
        \centering
        \includegraphics[width=\textwidth]{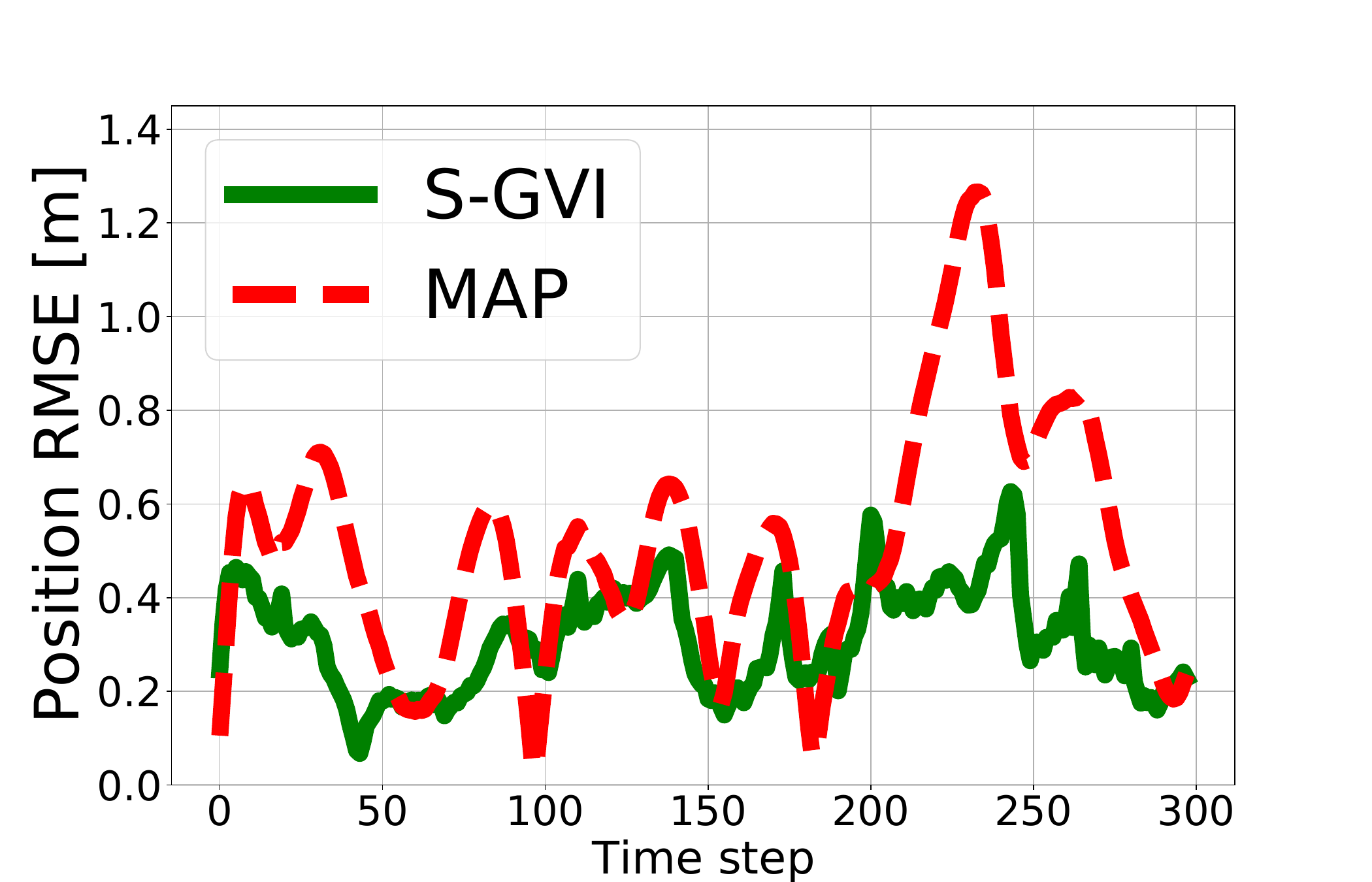}
        {{\scriptsize (d) Experiment 4}}
    \end{minipage}
    \caption{{\small Position RMSE plots of the proposed S-GVI and MAP estimation in each experiment. 
    }}
    \label{fig:RMSE_result}
\end{figure}

\begin{figure}[!t]
    \centering
    \begin{minipage}[t]{0.22\textwidth}
        \centering
        \includegraphics[width=\textwidth]{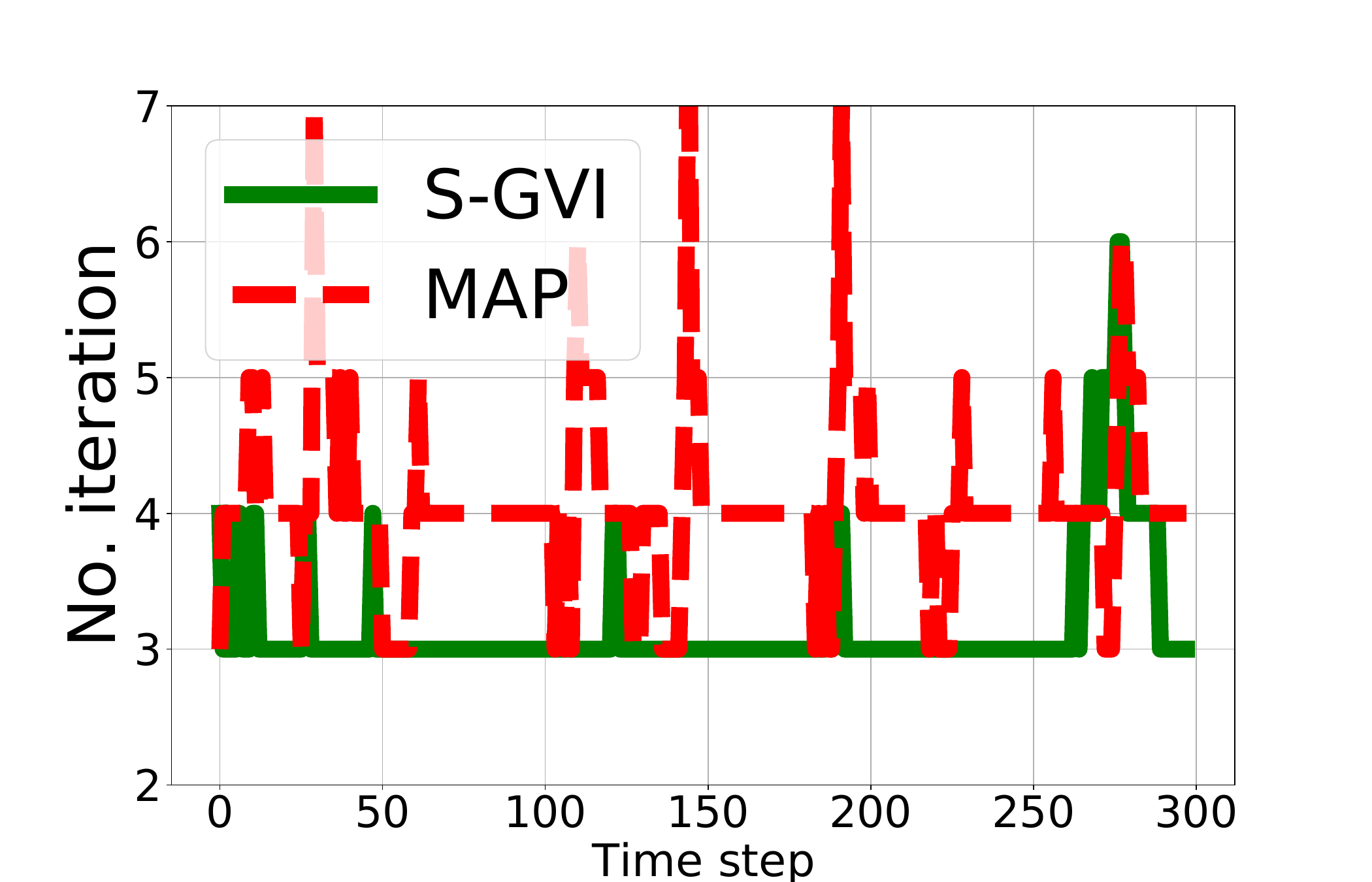}
        {{\scriptsize (a) Experiment 1}}
    \end{minipage}
    \hskip\baselineskip
    \begin{minipage}[t]{0.22\textwidth}
        \centering
        \includegraphics[width=\textwidth]{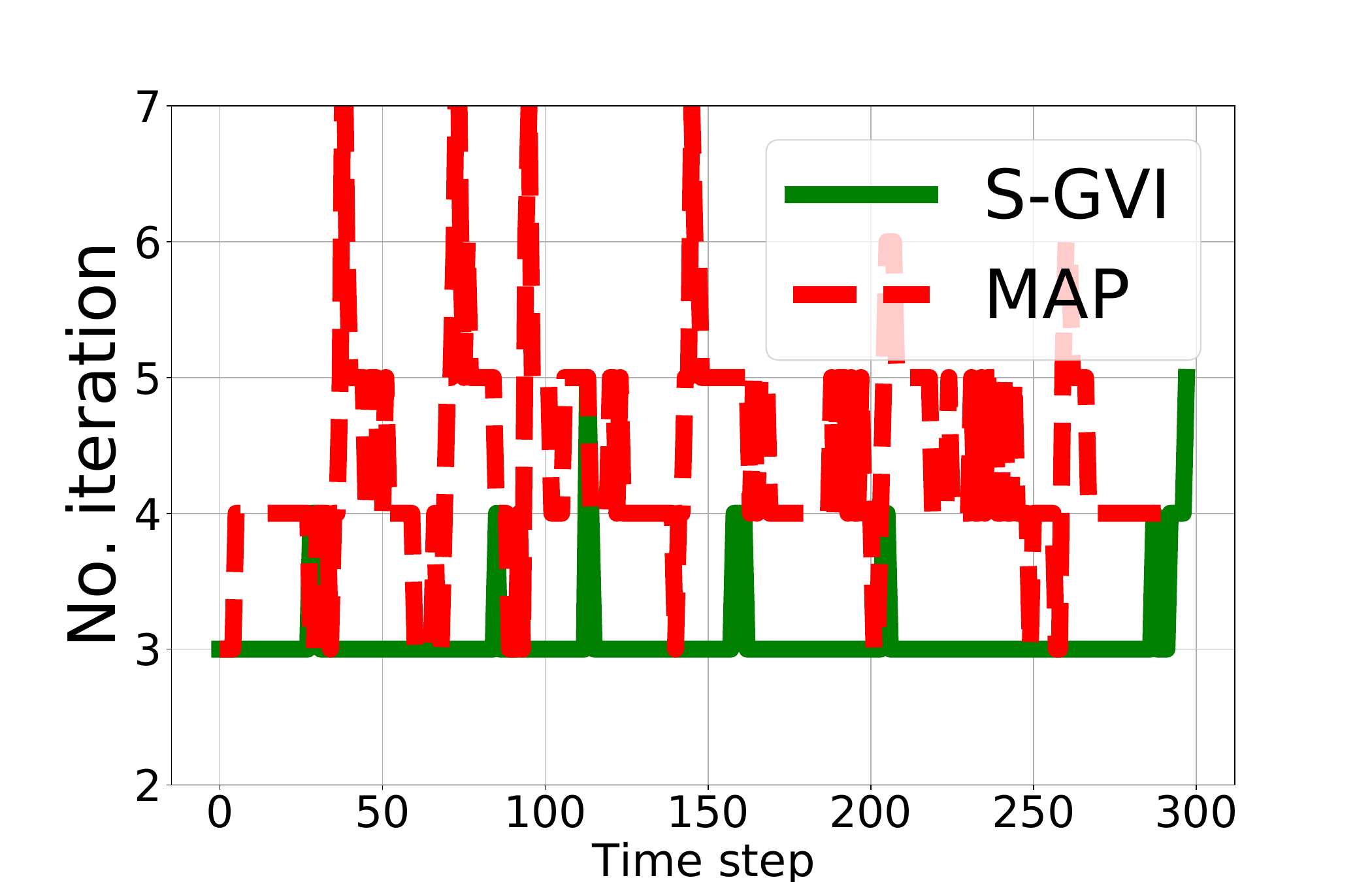}
        {{\scriptsize (b) Experiment 2}}
    \end{minipage}
    \hskip\baselineskip
    \begin{minipage}[t]{0.22\textwidth}
        \centering
        \includegraphics[width=\textwidth]{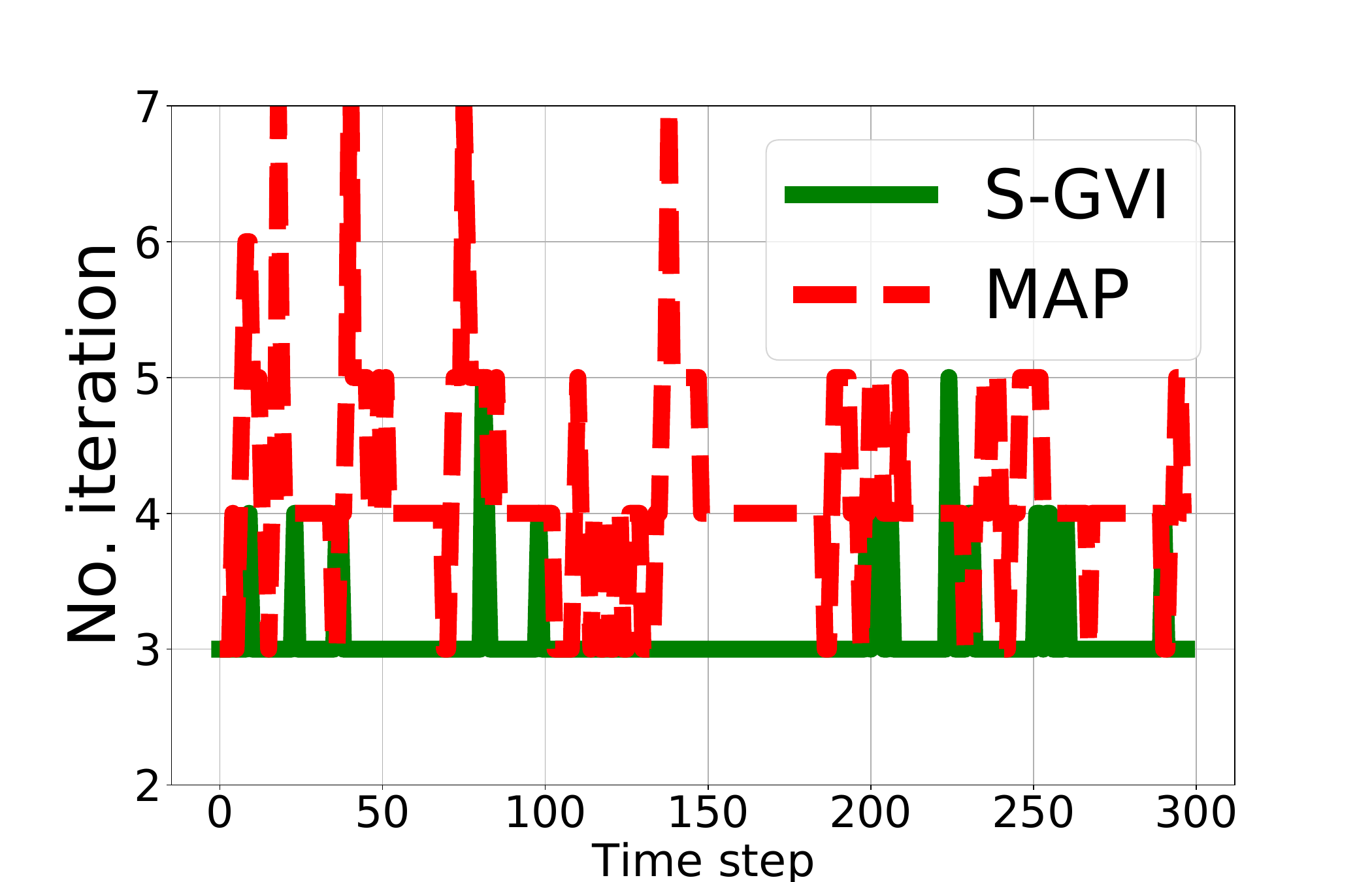}
        {{\scriptsize (c) Experiment 3}}
    \end{minipage}
    \hskip\baselineskip
    \begin{minipage}[t]{0.22\textwidth}
        \centering
        \includegraphics[width=\textwidth]{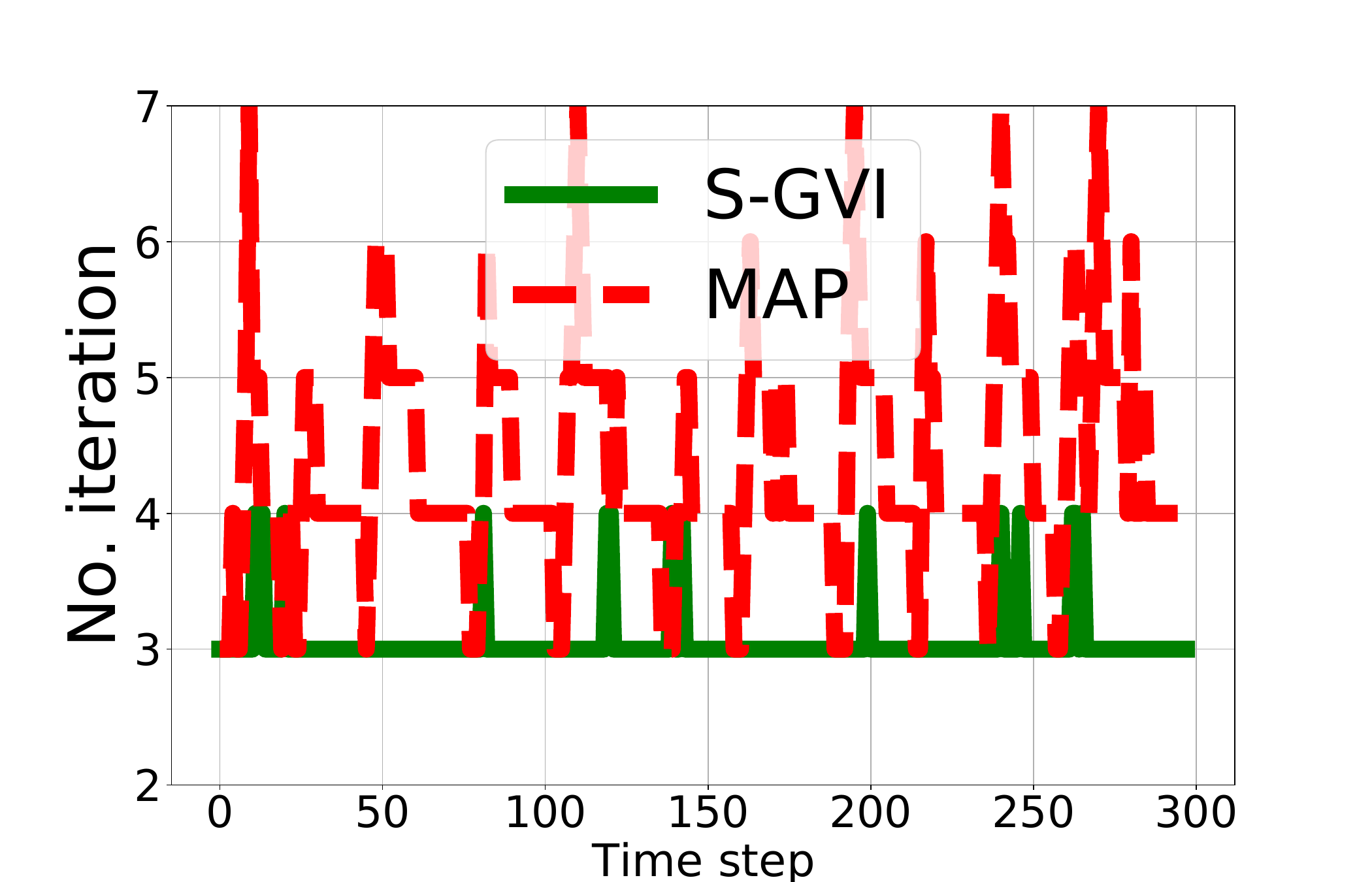}
        {{\scriptsize (d) Experiment 4}}
    \end{minipage}
    \caption{{\small Number of iterations for the proposed S-GVI and MAP estimations in each experiment. The S-GVI requires fewer iterations than MAP most of the time.}}
    \label{fig:Iter_result}
\end{figure}

The localization results in Fig.~\ref{fig:RMSE_result} and Table~\ref{Table_1} indicate that S-GVI consistently delivers superior performance, achieving lower RMSE compared to the MAP estimation. In the table, RMSE is averaged over the trajectory. Furthermore, since S-GVI accounts for the geometry of each parameter axis by using the FIM as an inner product in the update equations, it results in fewer iterations over most of the trajectory compared to MAP, as illustrated in Fig.~\ref{fig:Iter_result} and Table~\ref{Table_2}. Here, the number of iterations is averaged over the trajectory. However, unlike MAP estimation, which employs analytical linearization at the mean point and therefore does not use numerical methods for expectation terms, S-GVI applies a quadrature rule using sigma points. As a result, the computation time depends on the number of sigma-points $m$, rather than solely on the mean point. We measured the wall-clock time on the trajectory to assess real-time performance, as shown in Table~\ref{Table_2}. In our experiment, the number of states is $n_x=5$, resulting in $m=11$ sigma points. Consequently, the wall-clock time is higher than that of the MAP estimation. However, the measured time is approximately $0.5$ seconds over $300$ steps, with each step taking $0.0017$ seconds, which is significant compared to a time step of $0.1$ seconds.

\begin{figure*}[!t]
    \centering
    \begin{minipage}[t]{0.24\textwidth}
        \centering
        \includegraphics[width=\textwidth]{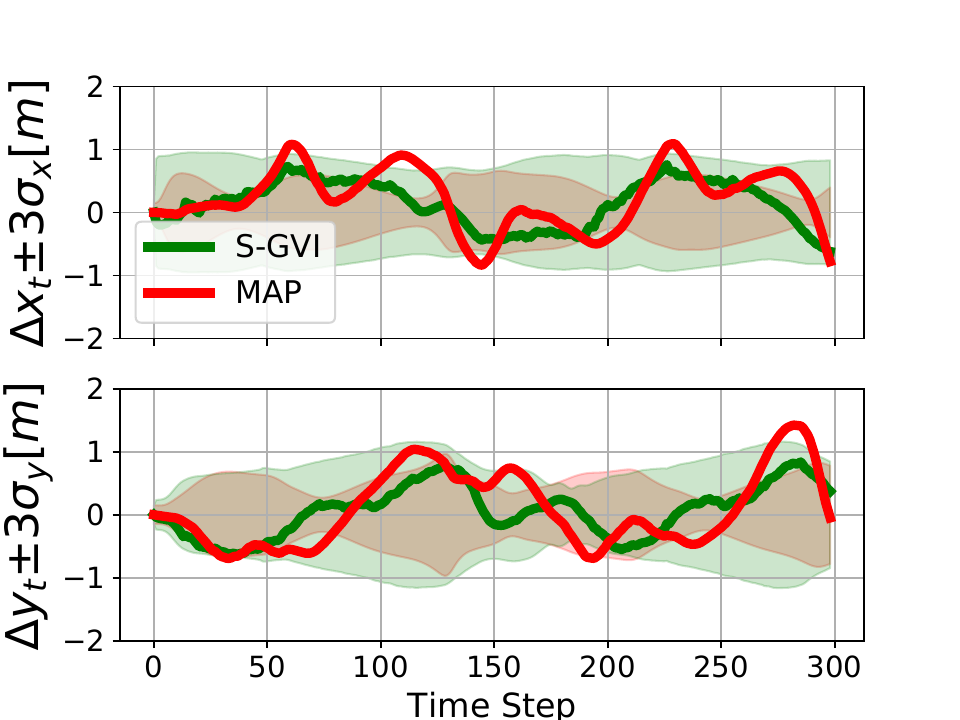}
        {{\scriptsize (a) Experiment 1}}
    \end{minipage}
    \begin{minipage}[t]{0.24\textwidth}
        \centering
        \includegraphics[width=\textwidth]{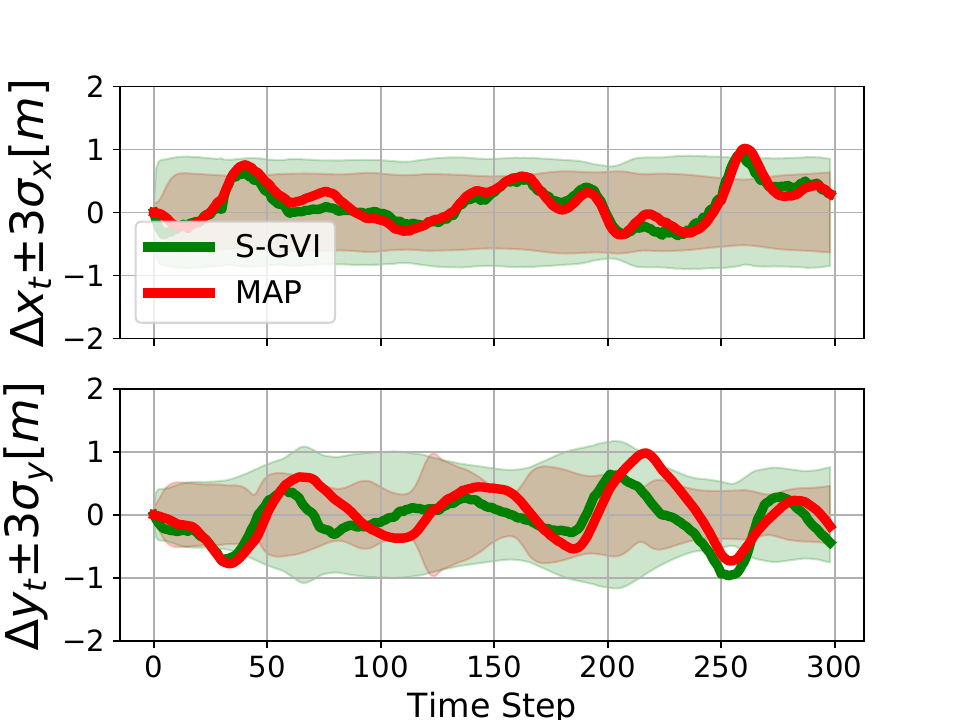}
        {{\scriptsize (b) Experiment 2}}
    \end{minipage}
    \begin{minipage}[t]{0.24\textwidth}
        \centering
        \includegraphics[width=\textwidth]{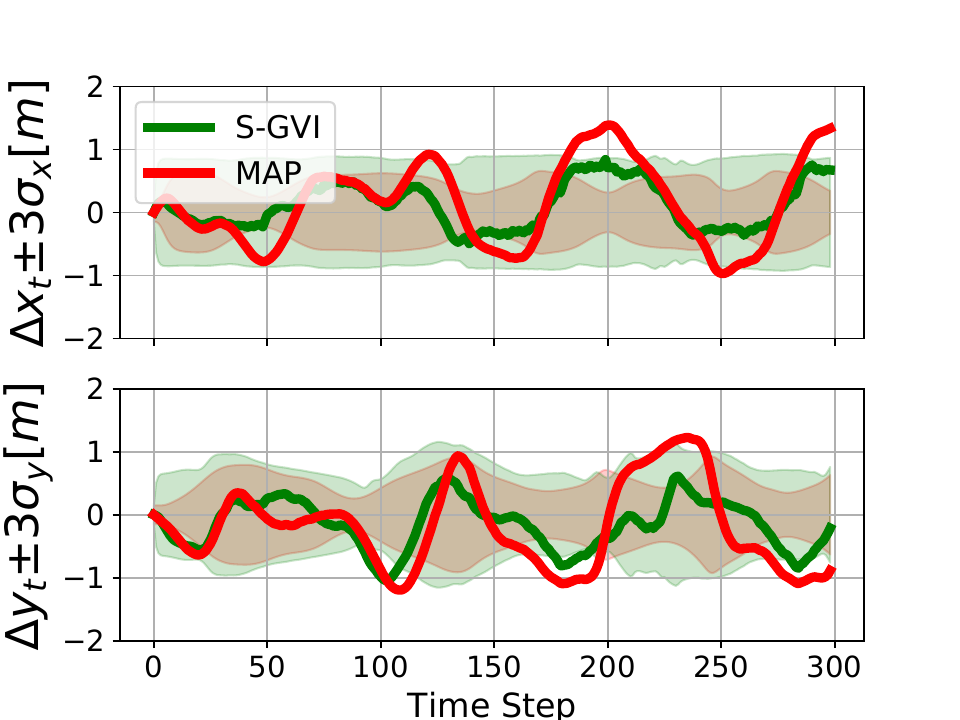}
        {{\scriptsize (c) Experiment 3}}
    \end{minipage}
    \begin{minipage}[t]{0.24\textwidth}
        \centering
        \includegraphics[width=\textwidth]{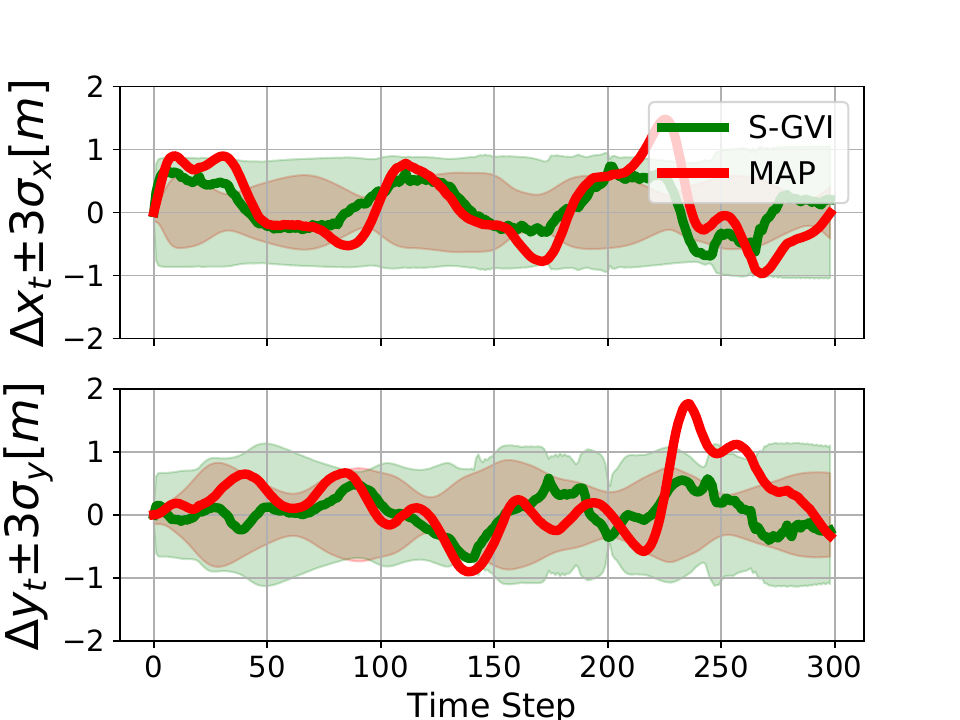}
        {{\scriptsize (d) Experiment 4}}
    \end{minipage}
    \caption{{\small Error with $3$-sigma bounds for the $x$ and $y$ positions of the proposed S-GVI and MAP estimation in each experiment.
    }}
    \label{fig:Exp_Result4}
\end{figure*}

We also evaluated the consistency of the estimation by comparing the error with the $3$-sigma bounds and the NEES score for the $x$ and $y$ positions in each experiment, as shown in Fig.~\ref{fig:Exp_Result4} and Table~\ref{Table_1}. In the MAP estimation, the estimated $x$ and $y$ positions exceeded the $3$-sigma error bounds (shaded region with light color) when the directional change was rapid, meaning the 2D estimated results lost consistency. In contrast, the S-GVI estimation briefly lost consistency, but it maintained it for most of the trajectory. Additionally, the averaged NEES score at the $5\%$ significance level for the S-GVI of the 2D position over the trajectory falls within the range $[0.051, 7.378]$, except for experiment $3$, indicating better consistency than the MAP estimation. For more details, see~\cite{bar2004estimation}. Furthermore, we observed that the MAP estimator diverged with changes in the initial covariance values $\Sigma_0$ and noise covariance values $Q$ and $R$. In contrast, the S-GVI is less sensitive to these parameters than the MAP estimation, leading to convergence in most cases. This divergence in the MAP estimator can be attributed to its methodology: it updates the mean and covariance in the correction step based on the result from the prediction step, making it susceptible to linearization errors, particularly when the coordinated turn model cannot fully capture the target motion. Conversely, S-GVI updates the mean and covariance in the posterior space, which considers the measurement data $z_t$, making it more robust against errors of the motion model.

Lastly, we compared the RMSE of SLR with that of the first-order Taylor expansion for marginalization in S-GVI. We replaced SLR with the Taylor expansion of $f$ at the mean point. The results in Table~\ref{Table_3} showed that SLR, which accounts for the uncertainty of linearization, captures nonlinearity better than the first-order Taylor expansion at the mean point.

\begin{table}[t]
\caption{Comparison of RMSE and NEES in Experiments}
\vspace{-0.1in}
\label{Table_1}
\begin{center}
\begin{tabular}{c | c c | c c }
\hline
& \multicolumn{2}{c|}{Experiment 1} & \multicolumn{2}{c}{Experiment 2}\\ 
& RMSE [m] & NEES & RMSE [m] & NEES \\ 
\hline
S-GVI & $0.384$ & $5.565$ & $0.297$ & $6.181$ \\ 
MAP & $0.501$ & $47.210$ & $0.351$ & $12.810$\\
\hline
& \multicolumn{2}{c|}{Experiment 3} & \multicolumn{2}{c}{Experiment 4}\\ 
& RMSE [m] & NEES & RMSE [m] & NEES\\ 
\hline
S-GVI & $0.366$ & $12.445$ & $0.312$ & $7.118$ \\ 
MAP & $0.634$ & $80.335$ & $0.514$ & $57.283$\\
\hline
\end{tabular}
\end{center}
\end{table}

\section{CONCLUSIONS} \label{sec::con}

We presented a novel Sequential Gaussian Variational Inference (S-GVI) approach for the nonlinear state estimation by leveraging sequential Bayesian principles within a Gaussian variational inference framework. Our results demonstrated that S-GVI with linear models reproduces the classical results of linear filtering. For nonlinear models, we proposed the use of statistical approximations and gradient optimization over the Gaussian space to address nonlinearity and enhance the efficiency of the optimization process in S-GVI. We showcased the robustness of S-GVI against nonlinearity through both simulation and real-world experiments. For higher-order dimensional systems, computational cost can be a prohibitive factor when using S-GVI in high-rate applications. Consequently, our future work will focus on developing efficient procedures for very high-dimensional systems. The source code is available on GitHub at \href{https://github.com/minwons/SGVI}{https://github.com/minwons/SGVI}.

\begin{table}[t]
\caption{Comparison of Averaged No. iteration and Wall-Clock Time}
\vspace{-0.1in}
\label{Table_2}
\setlength{\tabcolsep}{3pt}\renewcommand{\arraystretch}{1.0}
\begin{center}
\begin{tabular}{c | c c | c c }
\hline
& \multicolumn{2}{c|}{Experiment 1} & \multicolumn{2}{c}{Experiment 2}\\ 
& No. iteration & Wall-clock (s) & No. iteration & Wall-clock (s) \\ 
\hline
S-GVI & $3.164$ & $0.513$ & $3.070$ & $0.516$ \\ 
MAP & $4.063$ & $0.072$ & $4.345$ & $0.102$\\
\hline
& \multicolumn{2}{c|}{Experiment 3} & \multicolumn{2}{c}{Experiment 4}\\ 
& No. iteration & Wall-clock (s) & No. iteration & Wall-clock (s)\\ 
\hline
S-GVI & $3.080$ & $0.516$ & $3.050$ & $0.502$ \\ 
MAP & $4.130$ & $0.074$ & $4.342$ & $0.076$\\
\hline
\end{tabular}
\end{center}
\end{table}

\begin{table}[t]
\caption{Comparison of RMSE (m) for Marginalization Approximations}
\vspace{-0.1in}
\label{Table_3}
\begin{center}
\begin{tabular}{c | c | c | c | c }
\hline
& Exp 1 & Exp 2 & Exp 3 & Exp 4 \\ 
\hline
S-GVI (w/ SLR) & $0.413$ & $0.345$ & $0.416$ & $0.326$ \\ 
S-GVI (w/ Jacobian) & $0.635$ & $0.346$ & $ 0.743$ & $0.512$\\
MAP & $0.661$ & $0.362$ & $0.754$ & $0.527$\\
\hline
\end{tabular}
\end{center}
\end{table}

\bibliographystyle{ieeetr}
\bibliography{bib/references.bib}

\begin{thebibliography}{10}

\bibitem{sarkka2023bayesian}
S.~S{\"a}rkk{\"a} and L.~Svensson, {\em Bayesian filtering and smoothing}, vol.~17.
\newblock Cambridge university press, 2023.

\bibitem{mukadam2016gaussian}
M.~Mukadam, X.~Yan, and B.~Boots, ``Gaussian process motion planning,'' in {\em 2016 IEEE international conference on robotics and automation (ICRA)}, pp.~9--15, IEEE, 2016.

\bibitem{kim2021probabilistic}
D.~Kim, M.~Park, and Y.-L. Park, ``Probabilistic modeling and bayesian filtering for improved state estimation for soft robots,'' {\em IEEE Transactions on Robotics}, vol.~37, no.~5, pp.~1728--1741, 2021.

\bibitem{papoulis2002probability}
A.~Papoulis and S.~Unnikrishna~Pillai, {\em Probability, random variables and stochastic processes}.
\newblock Mc Graw Hill, 2002.

\bibitem{arasaratnam2007discrete}
I.~Arasaratnam, S.~Haykin, and R.~J. Elliott, ``Discrete-time nonlinear filtering algorithms using {G}auss--{H}ermite quadrature,'' {\em Proceedings of the IEEE}, vol.~95, no.~5, pp.~953--977, 2007.

\bibitem{svensen2007pattern}
M.~Svens{\'e}n and C.~M. Bishop, ``Pattern recognition and machine learning,'' 2007.

\bibitem{doucet2001introduction}
A.~Doucet, N.~De~Freitas, and N.~Gordon, ``An introduction to sequential {M}onte {C}arlo methods,'' {\em Sequential Monte Carlo methods in practice}, pp.~3--14, 2001.

\bibitem{gilks1995markov}
W.~R. Gilks, S.~Richardson, and D.~Spiegelhalter, {\em Markov chain Monte Carlo in practice}.
\newblock CRC press, 1995.

\bibitem{gelb1974applied}
A.~Gelb {\em et~al.}, {\em Applied optimal estimation}.
\newblock MIT press, 1974.

\bibitem{julier1997new}
S.~J. Julier and J.~K. Uhlmann, ``New extension of the {K}alman filter to nonlinear systems,'' in {\em Signal processing, sensor fusion, and target recognition VI}, vol.~3068, pp.~182--193, Spie, 1997.

\bibitem{kass1991laplace}
R.~E. Kass, L.~Tierney, and J.~B. Kadane, ``Laplace’s method in bayesian analysis,'' {\em Contemporary Mathematics}, vol.~115, pp.~89--99, 1991.

\bibitem{kullback1951information}
S.~Kullback and R.~A. Leibler, ``On information and sufficiency,'' {\em The annals of mathematical statistics}, vol.~22, no.~1, pp.~79--86, 1951.

\bibitem{blei2017variational}
D.~M. Blei, A.~Kucukelbir, and J.~D. McAuliffe, ``Variational inference: A review for statisticians,'' {\em Journal of the American statistical Association}, vol.~112, no.~518, pp.~859--877, 2017.

\bibitem{amari2016information}
S.-i. Amari, {\em Information geometry and its applications}, vol.~194.
\newblock Springer, 2016.

\bibitem{rudner2022tractable}
T.~G. Rudner, Z.~Chen, Y.~W. Teh, and Y.~Gal, ``Tractable function-space variational inference in bayesian neural networks,'' {\em Advances in Neural Information Processing Systems}, vol.~35, pp.~22686--22698, 2022.

\bibitem{fellows2019virel}
M.~Fellows, A.~Mahajan, T.~G. Rudner, and S.~Whiteson, ``Virel: A variational inference framework for reinforcement learning,'' {\em Advances in neural information processing systems}, vol.~32, 2019.

\bibitem{smidl2008variational}
V.~Smidl and A.~Quinn, ``Variational bayesian filtering,'' {\em IEEE Transactions on Signal Processing}, vol.~56, no.~10, pp.~5020--5030, 2008.

\bibitem{hu2018iterative}
Y.~Hu, X.~Wang, H.~Lan, Z.~Wang, B.~Moran, and Q.~Pan, ``An iterative nonlinear filter using variational bayesian optimization,'' {\em Sensors}, vol.~18, no.~12, p.~4222, 2018.

\bibitem{seo2023stein}
M.-W. Seo and S.~S. Kia, ``Stein-map: A sequential variational inference framework for maximum a posteriori estimation,'' {\em arXiv preprint arXiv:2312.08684}, 2023.

\bibitem{barfoot2020exactly}
T.~D. Barfoot, J.~R. Forbes, and D.~J. Yoon, ``Exactly sparse {G}aussian variational inference with application to derivative-free batch nonlinear state estimation,'' {\em The International Journal of Robotics Research}, vol.~39, no.~13, pp.~1473--1502, 2020.

\bibitem{courts2021gaussian}
J.~Courts, A.~G. Wills, and T.~B. Sch{\"o}n, ``Gaussian variational state estimation for nonlinear state-space models,'' {\em IEEE Transactions on Signal Processing}, vol.~69, pp.~5979--5993, 2021.

\bibitem{lambert2022continuous}
M.~Lambert, S.~Bonnabel, and F.~Bach, ``The continuous-discrete variational kalman filter (cd-vkf),'' in {\em 2022 IEEE 61st Conference on Decision and Control (CDC)}, pp.~6632--6639, IEEE, 2022.

\bibitem{yu2023gaussian}
H.~Yu and Y.~Chen, ``A {G}aussian variational inference approach to motion planning,'' {\em IEEE Robotics and Automation Letters}, vol.~8, no.~5, pp.~2518--2525, 2023.

\bibitem{wong2020variational}
J.~N. Wong, D.~J. Yoon, A.~P. Schoellig, and T.~D. Barfoot, ``Variational inference with parameter learning applied to vehicle trajectory estimation,'' {\em IEEE Robotics and Automation Letters}, vol.~5, no.~4, pp.~5291--5298, 2020.

\bibitem{goudar2022gaussian}
A.~Goudar, W.~Zhao, T.~D. Barfoot, and A.~P. Schoellig, ``Gaussian variational inference with covariance constraints applied to range-only localization,'' in {\em 2022 IEEE/RSJ International Conference on Intelligent Robots and Systems (IROS)}, pp.~2872--2879, IEEE, 2022.

\bibitem{dellaert2021factor}
F.~Dellaert, ``Factor graphs: Exploiting structure in robotics,'' {\em Annual Review of Control, Robotics, and Autonomous Systems}, vol.~4, pp.~141--166, 2021.

\bibitem{barfoot2024state}
T.~D. Barfoot, {\em State estimation for robotics}.
\newblock Cambridge University Press, 2024.

\bibitem{khan2017variational}
M.~E. Khan, W.~Lin, V.~Tangkaratt, Z.~Liu, and D.~Nielsen, ``Variational adaptive-newton method for explorative learning,'' {\em arXiv preprint arXiv:1711.05560}, 2017.

\bibitem{opper2009variational}
M.~Opper and C.~Archambeau, ``The variational {G}aussian approximation revisited,'' {\em Neural computation}, vol.~21, no.~3, pp.~786--792, 2009.

\bibitem{ver2012invented}
J.~M. Ver~Hoef, ``Who invented the delta method?,'' {\em The American Statistician}, vol.~66, no.~2, pp.~124--127, 2012.

\bibitem{ito2000gaussian}
K.~Ito and K.~Xiong, ``Gaussian filters for nonlinear filtering problems,'' {\em IEEE transactions on automatic control}, vol.~45, no.~5, pp.~910--927, 2000.

\bibitem{julier2004unscented}
S.~J. Julier and J.~K. Uhlmann, ``Unscented filtering and nonlinear estimation,'' {\em Proceedings of the IEEE}, vol.~92, no.~3, pp.~401--422, 2004.

\bibitem{schmidt2017minimizing}
M.~Schmidt, N.~Le~Roux, and F.~Bach, ``Minimizing finite sums with the stochastic average gradient,'' {\em Mathematical Programming}, vol.~162, pp.~83--112, 2017.

\bibitem{rostami2024forward}
M.~Rostami and S.~Kia, ``Forward gradient-based {F}rank-{W}olfe optimization for memory efficient deep neural network training,'' {\em arXiv preprint arXiv:2403.12511}, 2024.

\bibitem{horn2012matrix}
R.~A. Horn and C.~R. Johnson, {\em Matrix analysis}.
\newblock Cambridge university press, 2012.

\bibitem{garcia2016iterated}
{\'A}.~F. Garc{\'\i}a-Fern{\'a}ndez, L.~Svensson, and S.~S{\"a}rkk{\"a}, ``Iterated posterior linearization smoother,'' {\em IEEE Transactions on Automatic Control}, vol.~62, no.~4, pp.~2056--2063, 2016.

\bibitem{gustafsson1996best}
F.~Gustafsson and A.~J. Isaksson, ``Best choice of coordinate system for tracking coordinated turns,'' in {\em Proceedings of 35th IEEE Conference on Decision and Control}, vol.~3, pp.~3145--3150, IEEE, 1996.

\bibitem{bar2004estimation}
Y.~Bar-Shalom, X.~R. Li, and T.~Kirubarajan, {\em Estimation with applications to tracking and navigation: theory algorithms and software}.
\newblock John Wiley \& Sons, 2004.

\bibitem{simon2006optimal}
D.~Simon, ``Optimal state estimation: Kalman, h$\infty$, and nonlinear approaches. hoboken,'' {\em NJ: John Wiley and Sons}, vol.~10, p.~0470045345, 2006.

\end{thebibliography}


\section*{Appendix A:S-GVI applied to Linear Models}\label{sec::SGVI_Linear}
\noindent Let the system and the measurement models be linear~as~\cite{sarkka2023bayesian}: 
\begin{align} 
\label{eq::LSSM}
    x_t \sim \mathcal{N}(A_{t|t-1}x_{t-1},\, Q_{t-1}),~~z_t \sim \mathcal{N}(H_t x_t,\, R_t). 
\end{align}
For simplicity of notation, we omit $t$ and $t|t-1$ in $(A_{t|t-1},H_t,Q_{t-1},R_t)$. For this linear model, the expectation terms of the cost function~\eqref{eq::Optimization_cost_final} are calculated analytically as: 
\begin{align}\label{eq::Optimization_cost_linear}
    &\mathcal{L}(\theta_t)  = \frac{1}{2} \text{tr}\Bigl( R^{-1} \bigl( (z_t - H\mu_t)(z_t - H\mu_t)^{\top} + H\Sigma_tH^{\top}\bigl) \nonumber \\
    &+ \bigl(A\Sigma_{t-1}A^{\top} \!\!+ Q\bigl)^{-1}\! \bigl((\mu_t - A\mu_{t-1})(\mu_t - A\mu_{t-1})^{\top} \!\!+ \Sigma_t \bigl)\Bigl) \nonumber \\
    &+\frac{1}{2} \ln\bigl(|\Sigma_t^{-1}| \bigl)+\text{const},  
\end{align}
where $\text{const}=\frac{1}{2} \ln\bigl( (2\pi )^{n_x}|R| \bigl) +\frac{1}{2} \ln\bigl( (2\pi )^{n_x}|Q| \bigl) -\frac{n_x}{2} - \\ \frac{1}{2} \ln\bigl( (2\pi)^{n_x}\bigl)$, 
and $\text{tr}(\cdot)$ is the trace operator. The details of the derivation are provided in Appendix B.
A notable property of the cost function~\eqref{eq::Optimization_cost_linear} is that it decomposes into two separable additive components: one dependent on $\mu_t$ and the other on $\Sigma_t$. Consequently, the optimal solution can be computed efficiently by setting the first derivative of each separable component in \eqref{eq::Optimization_cost_linear} to zero and solving for $\mu_t$ and $\Sigma_t$, which gives us: 
\begin{subequations}
\label{eq::update_liner}
\begin{align}
      &\Sigma_t^{-1} = H^{\top}R^{-1}H + (A\Sigma_{t-1} A^{\top} + Q)^{-1},\\
      &\mu_t = \Sigma_t\bigl( H^{\top}R^{-1}z_t + (A\Sigma_{t-1} A^{\top} + Q)^{-1}A\mu_{t-1}  \bigl),
\end{align}
\end{subequations}
with details in Appendix B. In the appendix, we also verify that the Hessian matrices evaluated at stationary points~\eqref{eq::update_liner} are positive definite, confirming that the critical point~\eqref{eq::update_liner} is the global minimum. Importantly, the update equations~\eqref{eq::update_liner} match exactly those of \emph{information filter} (canonical form)~\cite{simon2006optimal}, demonstrating that S-GVI, when applied to linear models, reproduces classic linear results. This alignment underscores the validity of the technique before extending it to more complex nonlinear state estimation.

\section*{Appendix B}
For linear measurement model, the second term in~\eqref{eq::Optimization_cost_final}, by direct integration, equates to
\begin{align*}
    &\mathbb{E}_{q_{\theta_t}}[\ln{p(z_t|x_t)}]  = -\frac{1}{2} \ln\bigl( (2\pi)^{n_x}|R| \bigl) \\
    &~~~ -\frac{1}{2} \text{tr}\Bigl( R^{-1} \bigl((z_t - H\mu_t)(z_t - H\mu_t)^{\top} + H\Sigma_tH^{\top}\bigl)\Bigl).
\end{align*}
The third term in~\eqref{eq::Optimization_cost_final}, the inner expectation is addressed first. By Lemma A.2 and Lemma A.3 in~\cite{sarkka2023bayesian}, the inner expectation can be computed analytically as follows:
\begin{align*}
    &\mathbb{E}_{q_{\theta_{t\text{-}1}}}[p(x_t|x_{t-1})] \nonumber \\ 
    &=\! \int\! \mathcal{N}(x_t|Ax_{t-1},\, Q)~\mathcal{N}(x_{t-1}|\mu_{t-1},\,\Sigma_{t-1}) dx_{t-1} \nonumber \\
    &=\!\! \int\! \mathcal{N}\biggl(\!
        \begin{bmatrix}
        x_{t-1} \\
        x_t
        \end{bmatrix}
        | 
        \begin{bmatrix}
        \mu_{t-1} \\
        A\mu_{t-1}
        \end{bmatrix}
        ,
        \begin{bmatrix}
        \Sigma_{t-1} & \!\Sigma_{t-1}A^{\top} \\
        A\Sigma_{t-1} &\! A\Sigma_{t-1}A^{\top} + Q
        \end{bmatrix} \!\!
        \biggl) dx_{t-1} \nonumber \\
    &= \mathcal{N}(x_t|A\mu_{t-1},A\Sigma_{t-1}A^{\top} + Q).
\end{align*}
\noindent Then, the third term is
\begin{align*}
    &\mathbb{E}_{q_{\theta_t}}\bigr[\ln \mathbb{E}_{q_{\theta_{t\text{-}1}}}[p(x_t|x_{t-1})]\bigr] \nonumber \\ 
    &= -\frac{1}{2} \ln\bigl( (2\pi)^{n_x}|A\Sigma_{t-1}A^{\top} + Q| \bigl)  \\
    &~~~ -\frac{1}{2} \text{tr}\Bigl(\! (A\Sigma_{t-1}A^{\top} \!\!+\! Q)^{-1}\! \bigl((\mu_t \!-\! A\mu_{t-1})(\mu_t \!-\! A\mu_{t-1})\!^{\top} \!\!\!+\! \Sigma_t\bigl)\!\Bigl)\!. \nonumber
\end{align*}
\noindent Substituting for all elements, the cost function~\eqref{eq::Optimization_cost_final} in the case of linear systems becomes~\eqref{eq::Optimization_cost_linear}.

The first derivatives of the cost function~\eqref{eq::Optimization_cost_linear} w.r.t. $\Sigma_t$ and $\mu_t$ are
\begin{align*}
\!\!  \! \!\! \frac{\partial \mathcal{L}}{\partial \Sigma_t} \!&=\! \frac{1}{2} \text{tr}\Bigl(\! H^{\top}\! R^{-1} H \!+\! (A\Sigma_{t-1}A^{\top}+Q)^{-1}\! - \Sigma_t^{-1}\! \Bigl),\!\!
    \\
   \frac{\partial \mathcal{L}}{\partial \mu_t^{\top}} &= \text{tr}\Bigl(-H^{\top} R^{-1} (z_t - H\mu_t) \nonumber \\ 
   &\qquad~~~ + (A\Sigma_{t-1}A^{\top}+Q)^{-1}(\mu_t - A\mu_{t-1})\Bigl).
\end{align*}
By setting the first derivatives to zero and using the matrix property that if $M = 0$ then $\text{tr}(M) = 0$, the stationary points of the cost function are obtained as shown in~\eqref{eq::update_liner}.

The second derivative of the cost function~\eqref{eq::Optimization_cost_linear} w.r.t. $\mu_t$ is
\begin{align*}
   \frac{\partial^2 \mathcal{L}}{\partial \mu_t^{\top} \partial \mu_t} \!=\! \text{tr}\Bigl(H^{\top} R^{-1} H + (A\Sigma_{t-1}A^{\top}+Q)^{-1} \Bigl) > 0,\!\!
\end{align*}
which is positive definite since $\Sigma_{t-1}$, $R$, and $Q$ are positive definite matrices. Then, using the basic properties of the vectorization operator $\text{vec}(\cdot)$, and the Kronecker product $\otimes$, the second derivative of the cost function w.r.t. $\Sigma_t$ is
\begin{align*} 
    \frac{\partial (\partial \mathcal{L}/ \partial \Sigma_t)}{\partial \text{vec}(\Sigma_t)^\top} = \frac{1}{2} \text{tr}\Bigl(\Sigma_t^{-1} \otimes \Sigma_t^{-1} \Bigl) > 0,
\end{align*}
which is positive definite since $\Sigma_t$ is positive definite.


\addtolength{\textheight}{-12cm}   


\end{document}